\documentclass[11pt]{article}

\usepackage{color}
\usepackage{amsmath}
\usepackage{amssymb}
\usepackage{graphicx}
\usepackage{fullpage}
\usepackage{enumerate}
\usepackage[american]{babel}
\usepackage[authoryear,round,comma,compress]{natbib}
\usepackage{natbib}
\usepackage{setspace}
\usepackage{bbm}
\bibpunct{(}{)}{,}{a}{}{,}

\DeclareMathOperator*{\argmax}{arg\,max}
\DeclareMathOperator*{\argmin}{arg\,min}

\newcommand {\beq}{\begin{equation}}
\newcommand {\eeq}{\end{equation}}
\newcommand {\beqn}{\begin{equation*}}
\newcommand {\eeqn}{\end{equation*}}
\newcommand {\bear}{\begin{eqnarray}}
\newcommand {\eear}{\end{eqnarray}}
\newcommand {\bearn}{\begin{eqnarray*}}
\newcommand {\eearn}{\end{eqnarray*}}

\newtheorem{theorem}{Theorem}

\newtheorem{lemma}{Lemma}

\newtheorem{remark}{Remark}
\def\qed{ \ \vrule width.2cm height.2cm depth0cm\iffalse \smallskip\fi }

\normalsize
\begin{document}
\title{\vspace{-0.3cm}
Optimal Exploration-Exploitation in a Multi-Armed-Bandit Problem with Non-Stationary Rewards}
%Non-stationary Multi-Armed-Bandits: \\Strategies and Optimal Adaptation to Dynamic Benchmarks}
%Adapting to %Known and Unknown
%Reward Variation\\ in Stochastic Multi-Armed-Bandit Problems}
\author{
{\sf Omar Besbes}%\thanks{Graduate School of Business, e-mail: {\tt ob2105@columbia.edu}}
\\Columbia University\and
{\sf Yonatan Gur}%\thanks{Graduate School of Business, e-mail: {\tt ygur14@gsb.columbia.edu}}
\\Stanford University\and
{\sf Assaf Zeevi}\thanks{This work is supported by NSF grant 0964170 and BSF grant 2010466. An extended abstract of this work, including some preliminary results, appeared in \cite{Bes-Gur-Zee-Nips2014}. Correspondence: {\tt ob2105@gsb.columbia.edu}, {\tt ygur@stanford.edu}, {\tt assaf@gsb.columbia.edu}.}
\\Columbia University
}
\vspace{0.3cm}
\date{April 4, 2019\\ {\it To appear in Stochastic Systems, 2019}}
\maketitle
\setstretch{1.25}
\vspace{-0.2cm}
\begin{abstract}
\noindent In a multi-armed bandit problem a gambler needs to choose at each round one of K arms, each characterized by an unknown reward distribution. The objective is to maximize cumulative expected earnings over a planning horizon of length $T$, and performance is measured in terms of \emph{regret} relative to a (static) oracle that \emph{knows} the identity of the best arm a priori. This problem has been studied extensively when the reward distributions do not change over time, and uncertainty essentially amounts to identifying the optimal arm. We complement this literature by developing a flexible nonparametric model for temporal uncertainty in the rewards. The extent of temporal uncertainty is measured via the cumulative mean change of the rewards over the horizon, a metric we refer to as temporal variation, and regret is measured relative to a (dynamic) oracle that plays the \emph{pointwise} optimal action at each period. Assuming that nature can choose any sequence of mean rewards such that their temporal variation does not exceed $V$ (a temporal uncertainty budget), we characterize the complexity of this problem via the \emph{minimax regret} which depends on $V$ (the hardness of the problem) the horizon length $T$ and the number of arms $K$.
\vspace{0.2cm}

\end{abstract}

\newpage
%\doublespace
\setstretch{1.45}
\vspace{-0.15cm}
\section{Introduction}
\vspace{-0.0cm}
\textbf{Background and motivation.}  In the prototypical multi-armed bandit (MAB) problem a gambler needs to choose at each round of play, $t=1,\ldots,T$  one of $K$ arms  each characterized by an unknown reward distribution. Reward realizations are only observed when an arm is selected, and the gambler's objective is to maximize cumulative expected earnings over the planning horizon. To achieve this goal, the gambler needs to experiment with multiple actions (pulling arms) in an attempt to identify the optimal choice, while simultaneously taking advantage of the information available at each step of the game to optimize immediate rewards. This tradeoff between information acquisition via exploration (which is forward looking) and the exploitation of the latter for immediate reward optimization (which is more myopic in nature) is fundamental in many problem areas;
examples include  clinical trials (\citealt{Zelen}), strategic pricing (\citealt{BER-VAL1996}), investment in innovation (\citealt{BER-HEG2005}), packet routing (\citealt{Awerbuch-Kleinberg}), on-line auctions (\citealt{Kleinberg-Leighton}), assortment selection (\citealt{Car-Gal2007}), and on-line advertising (Pandey et al.  \citeyear{Pandey-etal}), to name but a few.
The broad applicability of this class of problems is one of the main reasons MAB problems have been so widely studied, since their inception in the seminal papers of \cite{Thom1933} and \cite{Robbins}.

In the MAB problem it has become commonplace to measure the performance of a policy relative to an oracle that {\it knows} the identity of the best arm a priori, with the gap between the two referred to as the {\it regret}.  If the regret is sublinear in $T$ this is tantamount to the policy being (asymptotically) long run average optimal, a first order measure of ``good'' performance. In their seminal paper \cite{Lai-Robbins} went further and identified a sharp characterization of the regret growth rate: no policy can achieve (uniformly) a regret that is smaller than order $\log T$ and there exists a class of policies, predicated on the concept of upper confidence bounds (UCB),  which achieve said growth rate of regret and hence are (second order) asymptotically optimal. The premultiplier in the growth rate of regret encodes the ``complexity'' of the MAB instance in terms of problem primitives, in essence, it is proportional to the number of arms $K$, and inversely proportional to a term that measure the ``distinguishability'' of the arms; roughly speaking, the closer the mean rewards are, the harder it is to differentiate the arms, and the larger this term is. When the aforementioned gap between the arms' mean reward can be arbitrarily small, the complexity of  the MAB problem, as measured by the growth rate of regret, is of order $\sqrt{T}$; see \cite{Auer-etal}. For overviews and further references the reader is referred to  the monographs by \cite{Berry-Fristedt}, \cite{Gittins-Book} for Bayesian / dynamic programming formulations, and \cite{Bub-Ces2012} that covers the machine learning literature and the so-called adversarial setting.

The parameter uncertainty in the MAB problem described above is purely {\it spatial} and the difficulty in the problem essentially amounts to uncovering the identity of the optimal arm with ``minimal'' exploration. However, in many application domains (including the ones mentioned above) {\it temporal changes} in the reward distribution structure are an intrinsic characteristic of the problem, and several attempts have been made to incorporate this into a stochastic MAB formulation.  The origin of this line of work can be traced back to \cite{Gittins-Jones} who considered a case where only the state of the chosen arm can change, giving rise to a rich line of work (see, e.g., \cite{Gittins}, and \cite{Whittle1}, as well as references therein). In particular, \cite{Whittle2} introduced the term \emph{restless bandits}; a model in which the states (associated with reward distributions) of arms change in each step according to an arbitrary, yet known, stochastic process. Considered a ``hard'' class of problems (cf. \citealt{Papa-Tsitsiklis}), this line of work has led to various approximations (see, e.g., \citealt{Bertsimas-Nino-Mora}), relaxations (see, e.g., \citealt{Guha-Munagala}), and restrictions of  the state transition mechanism (see, e.g., \cite{Ort-Rya-Auer-Mun2012} for irreducible Markov processes, and \cite{Azar-Laz-Bru2014} for a class of history-dependent rewards).

An alternative and more pessimistic approach views the MAB problem as a game between  the policy designer (gambler) and nature (adversary), where the latter can change the reward distribution of the arms at every instance of play. These ideas dates back to work of \cite{Bla1956} and \cite{Han1957} and has since seen significant development; \cite{Foster-Vohra},  \cite{Cesa-Bia-Lugosi}, and \cite{Bub-Ces2012} provide reviews of this line of research. Within this so called  {\it adversarial} formulation, the efficacy of a policy over a given time horizon $T$ is measured relative to a benchmark which is defined by the {\it single best action in hindsight}; the best action that could have been taken after seeing all reward realizations. The single best action represents a \emph{static oracle} and the regret in this formulation uses that as a benchmark.  For obvious reasons, this static oracle can perform quite poorly relative to a {\it dynamic oracle} that follows the dynamic optimal sequence of actions, as the latter optimizes the (expected) reward at each time instant.\footnote{Under non-stationary reward structure it is immediate that the single best action may be sub-optimal in a large number of decision epochs, and the gap between the performance of the static and the dynamic oracles can grow linearly with $T$.}  Thus, a potential limitation of the adversarial framework is that even if a policy exhibits a ``small'' regret relative to the static oracle, there is no guarantee it will perform well with respect to the more stringent dynamic oracle.

\paragraph*{Main contributions.} In this paper we provide a non-parametric formulation that is useful for modeling non-stationary rewards, allows to benchmark performance against a dynamic oracle, and yet is tractable from an analytical and computational standpoint, allowing for  a sharp characterization of problem complexity.  Specifically, our contributions are as follows.

\emph{Modeling.} We introduce a nonparametric modeling paradigm for non-stationary reward environments that we demonstrate to be tractable for analysis purposes, yet extremely flexible and broad in the scope of problem settings to which it can be applied. The key construct in this modeling framework is that of a budget of {\it temporal uncertainty}, which is measured in the total variation metric with respect to the cumulative mean reward changes over the horizon $T$. One can think of this as a temporal uncertainty set with a certain prescribed ``radius,''  $V_T$, such that all mean reward sequences that reside in this set have temporal variation that is less than this value. (These concepts are related to the ones  introduced in \cite{SO2013}  in the context of non-stationary stochastic approximation.) In particular, $V_T$ plays a central role in providing a sharp characterization of  the MAB problem complexity (in conjunction with the time horizon $T$ and the number of arms $K$). In this manner the paper advances our understanding of multi-armed bandit problems and complements existing approaches to modeling and analyzing non-stationary environments. In particular,  the nonparametric formulation we propose allows for very general temporal evolutions, extending most of the treatment in the non-stationary stochastic MAB literature which mainly focuses on a finite number of changes in the mean rewards, see, e.g., \cite{Gar-Mou2011}. Concomitantly,  as indicated above, the framework allows for the  more stringent dynamic oracle to be used as benchmark,  as opposed to the static benchmark used in the adversarial setting. (We further discuss these connections in \S\ref{sec:upp2}.)

\emph{Minimax regret characterization.} When the cumulative temporal variation  over the decision horizon $T$ is known to be bounded ex ante by $V_{T}$, we characterize the order of the minimax regret and hence the complexity of the learning problem.   It is worthwhile emphasizing that  the regret is measured here with respect to a \emph{dynamic oracle} that knows ex ante the mean reward evolution, and hence the \emph{dynamic sequence of best actions}. This is a marked departure from the weaker benchmark associated with the best single action in hindsight, which is typically used in the adversarial literature (exceptions noted below). First we establish lower bounds on the performance of {\it any} non-anticipating policy relative to the aforementioned  dynamic oracle, and then show that the order of this bound can be achieved, uniformly over the class of temporally varying reward sequences, by a suitably constructed  policy. In particular, the minimax regret is shown be of the order of $\left(KV_{T}\right)^{1/3} T^{2/3}$. The reader will note  that this result is quite distinct from the traditional stochastic MAB problem. In particular, in that problem the regret is either of order $K \log T$ (in the case of ``well separated'' mean rewards) or of order $K \sqrt{T}$ (the minimax formulation). In contrast, in  the non-stationary MAB problem the minimax complexity exhibits a very different dependence on problem primitives. In particular, even if the variation budget $V_T$ is a constant independent of $T$, then asymptotically the regret grows significantly faster, order $T^{2/3}$,  compared to the stationary case, and when $V_T$ itself grows with $T$, the problem exhibits complexity on a spectrum of scales.  Ultimately if the variation budget is such that $V_T$ is of the same order  as the time horizon $T$,  then the regret is linear (and no policy can achieve sublinear regret).

\textit{Elucidating exploration-exploitation tradeoffs in the presence of non-stationary rewards.} Unlike the traditional stochastic MAB problem, where the key trade-off is between the information acquired through exploration of the action space, and the immediate reward obtained by exploiting said information, the non-stationary MAB problem has further subtlety. In particular, while the policy we propose accounts for the explore-exploit tension, it highlights an additional consideration which concerns {\it memory}. More broadly, %unlike policies designed to perform well in stationary settings,
changes in the reward distribution that are inherent in the non-stationary stochastic setting require that a policy also ``forgets" the acquired information at a suitable rate.

\textit{Towards adaptive policies.} Our work provides a sharp characterization of the minimax complexity of the non-stationary MAB via a policy that knows a priori a bound, $V_T$,  on the mean reward temporal variation. This leaves open the question of online adaptation to said temporal variation. Namely, are there policies that are minimax- or nearly-minimax- optimal (in order) that do not require said knowledge, and hence can {\it adapt} to the nature of the changing mean reward sequences on the fly. This adaption means a policy can  achieve ex post performance which is as good (or nearly as good) as the one achievable under ex ante knowledge of the temporal variation budget. Section \ref{sec:uppadapting} in this paper lays out the key challenges associated with adapting to unknown variation budgets. While we are not able to provide an answer to the question,  we propose a potential solution methodology in the form of an  ``envelope" policy which employs several subordinate policies that each are constructed under a different assumption on the  temporal variation budget. The subordinate policies each represent a ``guess'' of the unknown temporal parameter, and the ``master'' envelope policy switches among these policies based on observed feedback, hence learning the changes in variation as they manifest. While we have no proof for optimality, or strong theoretical indication to believe an optimal adaptive policy is to be found in this family, numerical results indicate that such a conjecture is plausible. A full theoretical analysis of adaptive policies is left as a direction for future research.

\paragraph*{Further contrast with related work.} The two closest papers to ours are Auer et al. (\citeyear{Auer-etal}) that is couched  in the adversarial setting, and \cite{Gar-Mou2011} that pursues the stochastic setting. In both papers the non-stationary reward structure is constrained such that the identity of the best arm can change only a \emph{finite} number of times.  The regret in these instances is shown to be of order $\sqrt{T}$. Our analysis complements these results by treating a broader and more flexible class of temporal changes in the reward distributions. Our framework, that considers a dynamic oracle as benchmark, adds a further element that was so far absent from the adversarial formulation, and provides a much more realistic comparator against which to benchmark a policy. The concept of variation budget was advanced in \cite{SO2013} in the context of a non-stationary stochastic approximation problem, complementing a large body of work in the online convex optimization literature. The analogous relationship can be seen between the present paper and the work on adversarial MAB problems, though the techniques and analysis are quite distinct from \cite{SO2013} that deals with continuous (convex) optimization.

%Other papers have studied settings that are different then ours, and where non-stationary in reward/cost structure is modelled using other variation measures. For example,
\cite{Hazan-Kale2011} consider a non-stochastic MAB setting where regret is measured relative to the more traditional single best action in hindsight. The regret relative to the latter depends on a quadratic variation/spread measure of the cost vectors (relative to the empirical average vector of costs). \cite{Slivkins2014Contextual} considers a contextual bandit setting with a regret bound that depends, among other things, on a Lipschitz-type constant that limits reward differences over the joint space of arms and contexts. Other forms of variation are also considered in \cite{Jad-Rak-Sha-Sri2015} in an online convex optimization setting.

Some more recent papers have followed up on ideas presented in the current paper in the context of several non-stationary sequential optimization settings. These include MAB settings where additional structure is imposed on the non-stationarity of mean rewards (e.g., \citealt{Lev-Cra-Man2017}), as well as other sequential optimization settings (see, e.g., \cite{Wei-Hon-Lu2016} in an expert advise context). A few recent papers focus on the question of how to design policies that adapt to unknown variation budgets, see, for example, \cite{Kar-Ana2016} and \cite{Luo-Wei-Aga-Lan2018}, as well as \cite{Cao-Zhe-Kve-Xie2018} and \cite{Che-Sim-Zhu2018} in different MAB settings, and \cite{Zha-Yan-Zho2018} in a full information online convex optimization setting.

%Questions similar in spirit were recently pursued in the machine learning and sequential decision making literature; examples include \cite{bub-sliv2012}, \cite{Sel-Sli2014}, and \cite{Auer-Chiang2016} that present algorithms that achieve (near) rate optimal performance in both stochastic and adversarial MAB settings without prior knowledge on the nature of environment, and \cite{San-Neu-Laz2014}, that consider an adversarial online convex optimization setting and derive algorithms that are rate optimal regardless of whether the target function is weakly or strongly convex.

\paragraph*{Structure of the paper.} The next section introduces the formulation of a stochastic non-stationary MAB problem. In \S3 we characterize the minimax regret using lower and upper bounds, through exhibiting a family of adaptive policies that achieve rate optimal performance. \S\ref{subsec:num} provides numerical results. In \S\ref{sec:uppadapting} we lay out the challenges associated with adapting to unknown variation budget.
Proofs can be found in the Appendix.

\vspace{-0.0cm}
\section{Problem Formulation}\label{sec:form}\vspace{-0.0cm}

Let $\mathcal{K} = \left\{1, \ldots, K\right\}$ be a set of arms. Let $t= 1,2,\ldots$ denote the sequence of decision epochs, where at for each $t$ the decision-maker pulls one of the $K$ arms, and obtains a reward $X^{k}_{t}\in~\left[0,1\right]$, where $X^{k}_{t}$ is a random variable with expectation $\mu^{k}_t = \mathbb{E}\left[X^{k}_t\right]$. We denote the best possible expected reward at decision epoch $t$ by
\[
\mu^{\ast}_t \;=\; \max_{k\in\mathcal{K}}\left\{\mu^k_t\right\},\quad t=1,2,\ldots
\]

\paragraph* {Temporal variation in the expected rewards.} We assume the expected reward of each arm $\mu^{k}_t$ may change at any decision point. We denote by $\mu^{k}$ the sequence of expected rewards of arm $k$: $\mu^{k} =~\left\{ \mu^{k}_t \;:\; t=1,2,\ldots\right\}$. In addition, we denote by $\mu$ the sequence of vectors of all $K$ expected rewards: $\mu =~\left\{\mu^{k}\;:\; k=1,\ldots,K\right\}$. We assume that the expected reward of each arm can change an arbitrary number of times and  track the extent of (cumulative) {\it temporal variation} over a given horizon $T$ using the following metric:
\begin{equation}\label{eq:tv}
 {\mathcal V}(\mu;T) \::=\:  \sum_{t=1}^{T-1}\sup_{k\in \mathcal{K}}\left|\mu_{t}^{k} - \mu_{t+1}^{k}\right|, \quad T=2,3,\ldots
\end{equation}

\begin{figure}[!ht]
\centering
\includegraphics[height=1.9in]{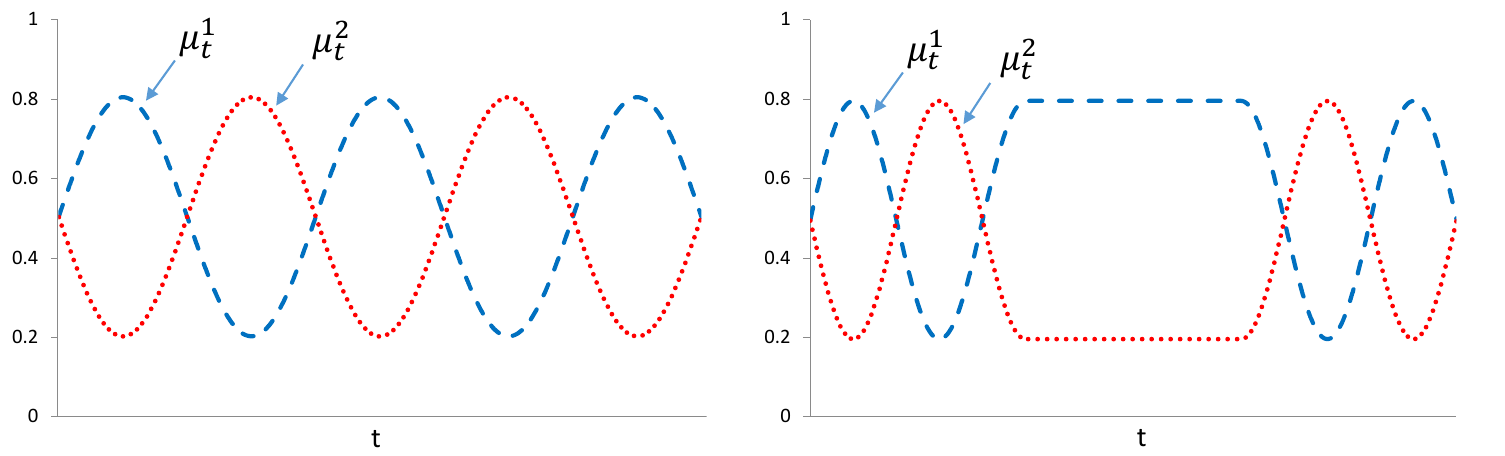}\vspace{-0.2cm}
\caption{\small Two instances of temporal changes  in the expected rewards of two arms that correspond to the same cumulative variation: ({\em Left\/}) Continuous variation in which a fixed variation (that equals 3) is spread over the whole horizon. ({\em Right\/}) A counterpart instance in which the same variation is ``spent" in the first and final thirds of the horizon, while mean rewards are fixed in between.
\label{fig:variation}}\vspace{-0.2cm}
\end{figure}

As will be further clarified later on, our formulation does not impose specific structure on $\mu$, but we assume that these are independent of the realized sample path of past actions. The formulation allows for many different forms in which the mean rewards may change; for illustration, Figure~\ref{fig:variation} depicts two different temporal patterns of mean reward changes that correspond to the same variation value ${\mathcal V}(\mu;T)$.\vspace{-0.4cm}

\noindent\textbf{Admissible policies, performance, and regret.} Let $U$ be a random variable defined over a probability space $\left(\mathbb{U}, \mathcal{U},\mathbf{P}_{u}\right)$. Let $\pi_{1}:\mathbb{U}\rightarrow \mathcal{K}$ and $\pi_{t}:[0,1]^{t-1}\times\mathbb{U}\rightarrow \mathcal{K}$ for $t=2,3,\ldots$ be measurable functions. With some abuse of notation we denote by $\pi_{t}\in \mathcal{K}$ the action at time $t$, that is given by\vspace{-0.2cm}
\begin{displaymath}
   \pi_{t} \;=\; \left\{
     \begin{array}{lr}
       \pi_{1}\left(U\right) & t=1,\quad\quad\quad\\
       \pi_{t}\left(X_{t-1}^{\pi}, \ldots, X_{1}^{\pi},U\right) & t=2,3,\ldots,\;
     \end{array}
   \right.\vspace{-0.2cm}
\end{displaymath}
The mappings $\left\{\pi_{t}:\;t=1,2,\ldots \right\}$ together with the distribution $\mathbf{P}_{u}$ define the class of admissible policies. We denote this class by $\mathcal{P}$. We further denote by $\left\{\mathcal{H}_{t},\;t=1,2,\ldots\right\}$ the filtration associated with a policy $\pi\in\mathcal{P}$, such that $\mathcal{H}_{1} = \sigma\left(U\right)$ and $\mathcal{H}_{t} = \sigma\left(\left\{X_{j}^{\pi}\right\}_{j=1}^{t-1},U\right)$ for all $t\in~\left\{2,3,\ldots\right\}$. Note that policies in $\mathcal{P}$ are non-anticipating, i.e., depend only on the past history of actions and observations, and allow for randomized strategies via their dependence on $U$.

For a given horizon $T$ and given sequence of mean reward vectors $\{\mu\}$ we define the \emph{regret} under policy $\pi\in \mathcal{P}$ compared to a \emph{dynamic} oracle as\vspace{-0.2cm}
\[
R^{\pi}(\mu,T) \;= \; \sum_{t=1}^{T}\mu^{\ast}_t - \mathbb{E}^{\pi}\left[\sum_{t=1}^{T}\mu^{\pi}_t\right],\vspace{-0.2cm}
\]
where the expectation $\mathbb{E}^{\pi}\left[\cdot\right]$ is taken with respect to the noisy rewards, as well as to the policy's actions. The regret measures the difference between the expected performance of the dynamic oracle that ``pulls'' the arm with highest mean reward at each epoch $t$, and that of any given policy. Note that in a stationary setting, one recovers the typical definition of regret where the oracle rule is constant.

\paragraph*{Budget of variation and minimax regret.} Let $\{V_t: t=1,2,\ldots\}$ be a non-decreasing sequence of positive real numbers such that $V_{1} = 0$, $KV_t \leq t$ for all $t$, and for normalization purposes set $V_2 = 2\cdot K^{-1}$. We refer to $V_T$ as the {\it variation budget} over time horizon $T$. For that horizon, we define the corresponding \emph{temporal uncertainty set} as the set of mean reward vector sequences with cumulative temporal variation that is bounded by the budget $V_T$,
\[
\mathcal{L}(V_T) \;=\; \left\{\mu \in \left[0,1\right]^{K\times T} \;:\;  {\mathcal V}(\mu;T) \leq V_{T}\right\}.\vspace{-0.1cm}
\]
The variation budget captures the constraint imposed on the non-stationary environment faced by the decision-maker. We denote by $\mathcal{R}^{\pi}(V_T,T)$ the regret guaranteed by policy $\pi$ {\it uniformly} over all mean rewards sequences $\{\mu\}$ residing in the temporal uncertainty set:
\[
\mathcal{R}^{\pi}(V_T,T) \;= \;\sup_{\mu\in \mathcal{L}(V_T)}R^{\pi}(\mu,T).\vspace{-0.1cm}
\]
In addition, we denote by $\mathcal{R}^{\ast}(V_T,T)$ the {\it minimax regret}, namely, the  minimal worst-case regret that can be guaranteed by an admissible policy $\pi\in\mathcal{P}$:
\[
\mathcal{R}^{\ast}(V_T,T)\; =\; \inf_{\pi\in \mathcal{P}} \sup_{\mu\in \mathcal{L}(V_T)}R^{\pi}(\mu,T).\vspace{-0.1cm}
\]
This minimax regret formulation implies that the sequence of mean rewards is selected by a non-adaptive adversary and is only constrained  to belonging the uncertainty set $\mathcal{L}(V_T)$.  Conditional on the sequence of mean rewards, our model is one of MAB with stochastic (noisy) rewards  that are generated by non-stationary distributions. A direct characterization of the minimax regret is hardly tractable. In what follows we will derive bounds  on the magnitude of this quantity as a function of the horizon $T$, that elucidate the impact of the budget of temporal variation $V_T$ on achievable performance; note that  $V_T$ may increase with the length of the horizon $T$.

\section{Analysis of the Minimax Regret}\label{sec:low}
\subsection{Lower bound}\label{sec:low}
We first provide a lower bound on the the best achievable performance.
\begin{theorem}\textbf{\textup{(Lower bound on achievable performance)}}\label{thm:low}
Assume that at each time $t=1,\ldots,T$ the rewards $X_t^k$, $k=1,\ldots,K$,  follow a Bernoulli distribution with mean $\mu_t^k$.  Then for any $T\geq 2$ and $V_{T}\in\left[K^{-1},K^{-1}T\right]$, the worst case regret for any policy $\pi\in\mathcal{P}$ is bounded below as follows:\vspace{-0.1cm}
\[
\mathcal{R}^{\pi}(V_T,T) \;\geq\; CK^{1/3} V_{T}^{1/3}T^{2/3},\vspace{-0.1cm}
\]
where $C$ is an absolute constant (independent of $T$ and $V_{T}$).
\end{theorem}

We note that when reward distributions are stationary and arm mean rewards can be arbitrarily close, there are known policies that achieve regret of order $\sqrt{T}$, up to logarithmic factors (cf.  \citealt*{Auer-etal2}). Moreover, it has been established in \cite*{Auer-etal} that this regret rate is still achievable even when there are changes in the mean rewards - as long as the number of changes is finite and independent of the horizon length $T$. Note that such sequences belong to  $\mathcal{L}(V_T)$ for the special case where $V_T$ is a constant independent of $T$. Theorem \ref{thm:low} states that when the notion of temporal variation is broader, as per the above, then it is no longer possible to achieve the aforementioned $\sqrt{T}$ performance. In particular, any policy must incur a regret of at least order $T^{2/3}$. At the extreme, when $V_T$ grows linearly with $T$, then it is no longer possible to achieve sublinear regret (and hence long run average optimality). The above theorem provides a full spectrum of bounds on achievable performance that range from $T^{2/3}$ to linear regret.\vspace{-0.3cm}

\begin{figure}[!t]
\centering
\includegraphics[height=1.6in]{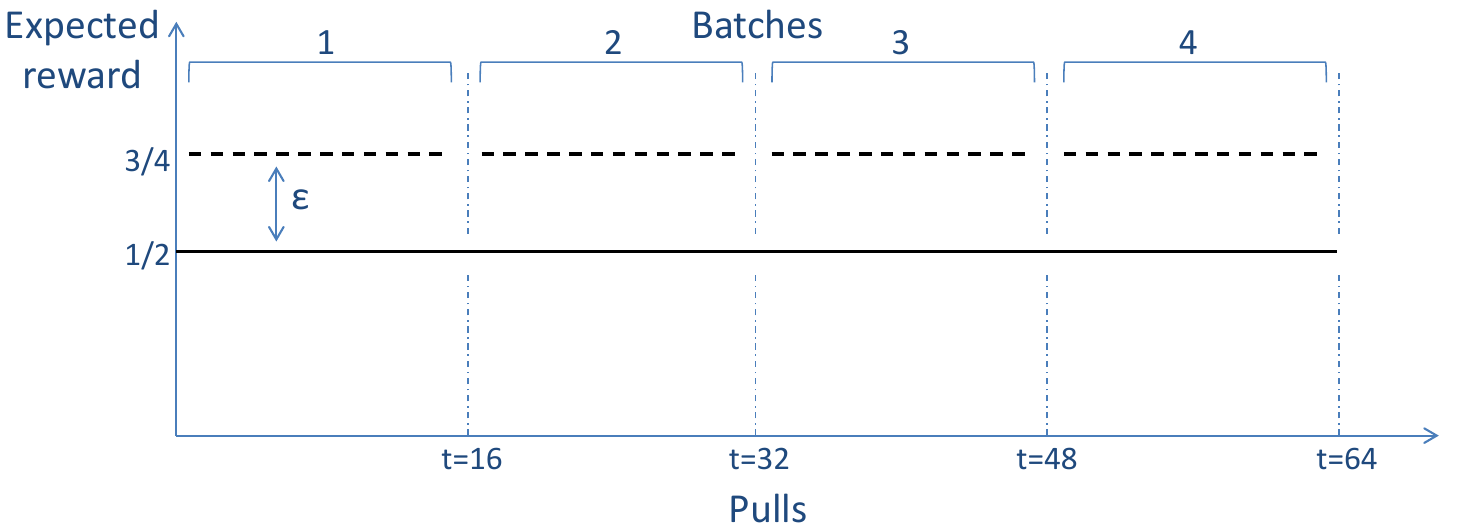}\vspace{-0.35cm}
\caption{\small Drawing a sequence from $\mathcal{L}'$. A numerical example of possible realizations of expected rewards. Here $T=64$, and we have $4$ decision batches, each contains $16$ pulls. We have $K^{4}$ equally probable realizations of reward sequences. In every batch one arm is randomly and independently drawn to have an expected reward of $1/2 +~\varepsilon$, where in this example $\varepsilon = 1/4$. This example corresponds to a variation budget of $V_{T}= ~\varepsilon\Delta =~1$.} \label{fig:lines}\vspace{-0.3cm}
\end{figure}

\paragraph*{Key ideas in the proof of Theorem \ref{thm:low}.} We define a subset of vector sequences $\mathcal{L}' \subset \mathcal{L}(V_T)$ and show that when $\mu$ is drawn randomly from $\mathcal{L}'$, any admissible policy must incur regret of order $V_{T}^{1/3}T^{2/3}$. We define a partition of the decision horizon $\mathcal{T} = \{1,\ldots,T\}$ into batches $\mathcal{T}_{1},\ldots,\mathcal{T}_{m}$ of size $\Delta$ each (except, possibly the last batch). In $\mathcal{L}'$, every batch contains exactly one ``good" arm with expected reward $1/2+\varepsilon$ for some $0 < \varepsilon \leq 1/4$, and all the other arms have expected reward $1/2$. The ``good" arm is drawn independently in the beginning of each batch according to a discrete uniform distribution over $\mathcal{K}$. Thus, the identity of the ``good" arm can change only between batches. See Figure \ref{fig:lines} for an example of possible realizations of a sequence $\mu$ that is randomly drawn from $\mathcal{L}'$.
By selecting $\varepsilon$ such that $\varepsilon T/ \Delta\leq V_{T}$, any $\mu\in\mathcal{L}'$ is composed of expected reward sequences with a variation of at most $V_{T}$, and therefore $\mathcal{L}' \subset \mathcal{L}(V_T)$. Given the draws under which expected reward sequences are generated, nature prevents any accumulation of information from one batch to another, since at the beginning of each batch a new ``good" arm is drawn independently of the history.

Using ideas from the analysis of \cite[proof of Theorem 5.1]{Auer-etal}, we establish that no admissible policy can identify the ``good" arm with high probability within a batch. Since there are $\Delta$ epochs in each batch, the regret that any policy must incur along a batch is of order $\Delta\cdot\varepsilon \approx~\sqrt{\Delta}$, which yields a regret of order $\sqrt{\Delta}\cdot T/\Delta \approx~T/\sqrt{\Delta}$ throughout the whole horizon. Theorem \ref{thm:low} then follows from selecting $\Delta \approx \left(T/V_{T}\right)^{2/3}$, the smallest feasible $\Delta$ that satisfies the variation budget constraint, yielding regret of order $V_{T}^{1/3}T^{2/3}$. \qed

\vspace{-0.1cm}
\subsection{Rate optimal policy}\label{sec:upp2}\vspace{-0.2cm}

We will now show that the order of the bound in Theorem \ref{thm:low} can be achieved. To that end, introduce the following policy, referred to as Rexp3.
\begin{center}\vspace{-6.0mm}
\line(1,0){490}
\end{center}
\vspace{-6.0mm}\textbf{Rexp3.} Inputs: a positive number $\gamma$, and a batch size $\Delta$.\vspace{-0.3cm}
\begin{enumerate}
  \item Set batch index $j = 1$\vspace{-0.3cm}
  \item Repeat while $j \leq \left\lceil T/\Delta\right\rceil$:\vspace{-0.2cm}
  \begin{enumerate}
    \item Set $\tau = \left(j-1\right)\Delta$\vspace{-0.1cm}
    \item Initialization: for any $k\in \mathcal{K}$ set $w^{k}_{t} = 1$\vspace{-0.1cm}
    \item Repeat for $t = \tau + 1,\ldots,\min\left\{T,\tau + \Delta\right\}$:\vspace{-0.1cm}
  \begin{itemize}
    \item For each $k\in\mathcal{K}$, set\vspace{-0.4cm}
    \[
    p^{k}_{t} \;=\; \left(1-\gamma\right)\frac{w^{k}_{t}}{\sum_{k'=1}^{K}w^{k'}_{t}} + \frac{\gamma}{K}\vspace{-0.1cm}
    \]
    \item Draw an arm $k'$ from $\mathcal{K}$ according to the distribution $\left\{p^{k}_{t}\right\}_{k=1}^{K}$, and receive a reward $X^{k'}_{t}$\vspace{-0.1cm}
    \item For $k'$ set $\hat{X}^{k'}_{t} = X^{k'}_{t}/p^{k'}_{t}$, and for any $k\neq k'$ set $\hat{X}^{k}_{t} = 0$. For all $k\in \mathcal{K}$ update:\vspace{-0.15cm}
        \[
        w^{k}_{t+1}\; =\; w^{k}_{t}\exp\left\{ \frac{\gamma \hat{X}^{k}_{t}}{K}  \right\}\vspace{-0.3cm}
        \]
  \end{itemize}
    \item Set $j=j+1$, and return to the beginning of step 2\vspace{-7.5mm}
  \end{enumerate}
\end{enumerate}
\begin{center}
\line(1,0){490}
\end{center}\vspace{-4mm}
Clearly $\pi\in \mathcal{P}$. The Rexp3 policy uses Exp3, a policy introduced by \cite{Fre-Sch1997} for solving a worst-case sequential allocation problem, as a subroutine, restarting it every $\Delta$ epochs. At each epoch, with probability $\gamma$ the policy explores by sampling an arm from a discrete uniform distribution over the set $\mathcal{K}$. With probability $\left(1-\gamma\right)$, the policy exploits information gathered thus far by drawing an arm according to weights that are updated exponentially based on observed rewards. Therefore $\pi$ balances exploration and exploitation by a proper selection of an \emph{exploration rate} $\gamma$ and a batch size $\Delta$. At a high level, the sampling probability $\left\{p^{k}_{t}\right\}$ represents the current ``certainty'' of the policy regarding the identity of the best arm.

\begin{theorem}\textbf{\textup{(Rate optimality)}}\label{thm:resregret}
Let $\pi$ be the Rexp3 policy with a batch size $\Delta =~\left\lceil\left(K\log K\right)^{1/3}\left(T/V_{T}\right)^{2/3}\right\rceil$ and with $\gamma = \min\left\{1 \;,\; \sqrt{\frac{K\log K}{(e-1)\Delta}} \right\}$. Then, for every $T\geq 2$, and $V_{T}\in\left[K^{-1}, K^{-1}T\right]$ the worst case regret for this policy is bounded from above as follows
\[
\mathcal{R}^{\pi}(V_T,T) \;\leq\; \bar{C}(K \log K)^{1/3} V_{T}^{1/3}T^{2/3},
\]
where $\bar{C}$ is an absolute constant independent of $T$ and $V_{T}$.
\end{theorem}

\iffalse
Theorem \ref{thm:resregret} is obtained by establishing a connection between the regret relative to the single best action, and the regret relative to the dynamic oracle in a non-stationary stochastic setting with a variation budget. Several classes of policies, such as exponential-weight (including Exp3) and polynomial-weight policies, have been shown to achieve regret of order $\sqrt{T}$ relative to the single best action in the adversarial setting (for an overview see chapter $6$ of \citealt{Cesa-Bia-Lugosi}). While in general these policies tend to perform well numerically, there is no guarantee for their performance relative to the \emph{dynamic oracle} studied here, since any static action may incur linear regret relative to the dynamic oracle; see also \cite{Har-etal2006} for a study of the empirical performance of one class of algorithms. The proof of Theorem~\ref{thm:resregret} actually shows that \emph{any} policy that achieves regret of order $\sqrt{T}$ relative to the single best action in the adversarial setting, can be used as a subroutine to obtain rate optimal performance in our setting.
\fi

The above theorem (in conjunction with the lower bound in Theorem \ref{thm:low}) establishes the order of the minimax regret, namely,
$$
 {\cal R}^*(V_T,T) \asymp (V_T)^{1/3} T^{2/3}.\vspace{0.1cm}
$$

\begin{remark}{\textbf{\textup{(Dependence on the number of arms)}}}
{\rm Our proposed policy, Rexp3, is driven primarily by its simplicity and the ability to elucidate key trade-offs in exploration-exploitation in the non-stationary problem setting. We note that there is a minor gap between  the lower and upper bounds insofar as their dependence on $K$ is concerned. In particular, the logarithmic term in $K$ in the upper bound  on the minimax regret can be removed by adapting other known policies that are designed for adversarial settings, for example, the INF policy that is described in \cite{Aud-Bub2009}.}
\end{remark}\vspace{0.1cm}

\subsection{Further extensions and discussion} \label{sec:upp3}

\paragraph*{A continuous update near-optimal policy.} The Rexp3 policy described above is particularly simple in  its structure and lends itself to elucidating the exploration-exploitation tradeoffs that exist in the non-stationary stochastic model.  It is, however, somewhat clunky in the manner in which it addresses the ``remembering-forgetting'' balance via the batching and abrupt restarts.   The following policy, introduced in \cite{Auer-etal}, can be modified to present a ``smoother'' counterpart to Rexp3.
\begin{center}\vspace{-0.45cm}
\line(1,0){490}
\end{center}
\vspace{-4.5mm}\textbf{Exp3.S.} Inputs: positive numbers $\gamma$ and $\alpha$.\vspace{-0.2cm}
\begin{enumerate}
    \item Initialization: for $t=1$, for any $k\in \mathcal{K}$ set $w^{k}_{t} = 1$\vspace{-0.2cm}
    \item For each $t = 1,2,\ldots$:\vspace{-0.2cm}
  \begin{itemize}
    \item For each $k\in\mathcal{K}$, set\vspace{-0.2cm}
    \[
    p^{k}_{t} \;=\; \left(1-\gamma\right)\frac{w^{k}_{t}}{\sum_{k'=1}^{K}w^{k'}_{t}} + \frac{\gamma}{K}\vspace{-0.1cm}
    \]
    \item Draw an arm $k'$ from $\mathcal{K}$ according to the distribution $\left\{p^{k}_{t}\right\}_{k=1}^{K}$, and receive a reward $X^{k'}_{t}$\vspace{-0.1cm}
    \item For $k'$ set $\hat{X}^{k'}_{t} = X^{k'}_{t}/p^{k'}_{t}$, and for any $k\neq k'$ set $\hat{X}^{k}_{t} = 0$\vspace{-0.1cm}
    \item For all $k\in \mathcal{K}$ update:\vspace{-2mm}
        \[
        w^{k}_{t+1}\; =\; w^{k}_{t}\exp\left\{ \frac{\gamma \hat{X}^{k}_{t}}{K}  \right\} + \frac{e\alpha}{K}\sum_{k'=1}^{K}w^{k'}_{t}\vspace{-2.5mm}
        \]
  \end{itemize}
\end{enumerate}
\begin{center}
\line(1,0){490}
\end{center}\vspace{-5.5mm}
The performance of Exp3.S was analyzed in \cite{Auer-etal} under a variant of the adversarial MAB formulation where the \emph{number of switches} $s$ in the identity of the best arm is finite. The next result shows that by selecting appropriate tuning parameters, Exp3.S guarantees near optimal performance in the much broader non-stationary stochastic setting we consider here.

\begin{theorem}\textbf{\textup{(Near-rate optimality for continuous updating)}}\label{thm:contregret}
Let $\pi$ be the Exp3.S policy with the parameters $\alpha = \frac{1}{T}$ and $\gamma = \min\left\{1, \left(\frac{4V_{T}K\log\left(KT\right)}{(e-1)^{2}T}\right)^{1/3}\right\}$. Then, for every $T\geq 2$ and $V_{T}~\in~\left[K^{-1}, K^{-1}T\right]$, the worst case regret is bounded from above as follows\vspace{-0.25cm}
\[
\mathcal{R}^{\pi}(\mathcal{V},T) \;\leq\; \bar{C}\left(K\log K\right)^{1/3}\left(V_{T}\log T \right)^{1/3}\cdot T^{2/3},\vspace{-0.25cm}
\]
where  $\bar{C}$ is an absolute constant  independent of $T$ and $V_{T}$.
\end{theorem}

\paragraph*{Minimax regret and relation to traditional (stationary) MAB problems.}  The minimax regret should be contrasted with the stationary MAB problem where $\sqrt{T}$ is the order of the minimax regret (see \cite{Auer-etal2}); if  the arms are well  separated then the order is $\log T$ (see \cite{Lai-Robbins}). To illustrate the type of regret performance driven by  the non-stationary environment, consider the case $V_{T} = C\cdot T^{\beta}$ for some $C>0$ and $0\leq \beta < 1$, where the minimax regret is of order $T^{\left(2 + \beta\right)/3}$.    The driver of the change is the optimal exploration-exploitation balance. Beyond the ``classical'' exploration-exploitation trade-off, an additional key element of our problem is the non-stationary nature of the environment. In this context, there is an additional tension  between ``remembering" and ``forgetting." Specifically, keeping track of more observations may decrease the variance of the  mean reward estimates, but ``older" information is potentially less useful and might bias our estimates. (See also discussion along these lines for UCB-type policies, though in a different set up,  in \cite{Gar-Mou2011}.)
 The design of Rexp3 reflects these considerations.  An exploration rate $\gamma \approx \Delta^{-1/2} \approx \left(V_{T}/T\right)^{-1/3}$ leads to an order of $\gamma T \approx V_{T}^{1/3}T^{2/3}$ exploration periods; significantly more than the order of $T^{1/2}$ explorations that is optimal in the stationary setting (with non-separated rewards).

\paragraph*{Relation to other non-stationary MAB instances.} The class of MAB problems with non-stationary rewards formulated in the current paper extends other MAB formulations that allow rewards to change in a more restricted manner. As mentioned earlier, when the variation budget grows linearly with the time horizon, the regret must grow linearly. This also implies the observation of \cite{Slivkins-Upfal} in a setting in which rewards evolve according to a Brownian motion and hence the regret is linear in $T$. Our results can also be positioned relative to those of \cite{Gar-Mou2011}, that study a stochastic MAB problems in which expected rewards may change a finite number of times, and \cite{Auer-etal} that formulate an adversarial MAB problem in which the identity of the best arm may change a finite number of times (independent of $T$). Both studies suggest policies that, utilizing  prior knowledge that the number of changes must be finite, achieve regret of order $\sqrt{T}$ relative to the best sequence of actions. As noted earlier, in our problem formulation this set of reward sequence instances would fall into the case where the variation budget $V_{T}$ is fixed and independent of $T$. In that case, our results establish that  the regret must be at least of  order $T^{2/3}$. Moreover, a careful look at the proof of Theorem \ref{thm:low} clearly identifies the ``hard'' set of sequences as those that have a ``large'' number of changes in the expected rewards, hence complexity in our problem setting is markedly different than the aforementioned studies. It is also worthwhile to note that  in \cite{Auer-etal}, the performance of Exp3.S is measured relative to a benchmark that resembles the dynamic oracle discussed in the current paper, but while the latter can switch arms in every epoch, the benchmark in \cite{Auer-etal} is limited to be a sequence of $s+1$ actions (each of them is ex-post optimal in a different segment of the decision horizon). This represents an extension of the ``single-best-action-in-hindsight'' benchmark, but is still far more restrictive than the dynamic oracle formulation we develop and pursue in the present paper.

\vspace{-0.0cm}
\section{Numerical Results}\label{subsec:num}\vspace{-0.0cm}
We illustrate our main results for the near-optimal policy with continuous updating detailed in section \ref{sec:upp3}. This policy, unlike Rexp3, exhibits much ``smoother'' performance as is more conducive for illustrative purposes.

\paragraph*{Setup.} We consider instances where two arms are available: $\mathcal{K} = \left\{1,2\right\}$. The reward $X^{k}_{t}$ associated with arm $k$ at epoch $t$ has a Bernoulli distribution with a changing expectation $\mu_{t}^{k}$:\vspace{-0.1cm}
\begin{displaymath}
   X^{k}_{t} = \left\{
     \begin{array}{lr}
       1 & \text{ w.p. } \mu^{k}_{t} \\
       0 & \text{ w.p. } 1 - \mu^{k}_{t}
     \end{array}
   \right.\vspace{-0.1cm}
\end{displaymath}
for all $t \in {\cal T} = \{ 1,\ldots,T\}$, and for any arm $k\in\mathcal{K}$. The evolution patterns of $\mu_{t}^{k}$, $k\in \mathcal{K}$ will be specified below. At each epoch $t\in \mathcal{T}$ the policy selects an arm $k\in \mathcal{K}$. Then, the reward $X_{t}^{k}$ is realized and observed. The mean loss at epoch $t$ is $\mu_{t}^{\ast} - \mu_{t}^{\pi}$. Summing over the horizon and replicating, the average of the empirical cumulative mean loss approximates the expected regret compared to the dynamic oracle.

\begin{figure}[t]
\centering
\includegraphics[height=2.0in]{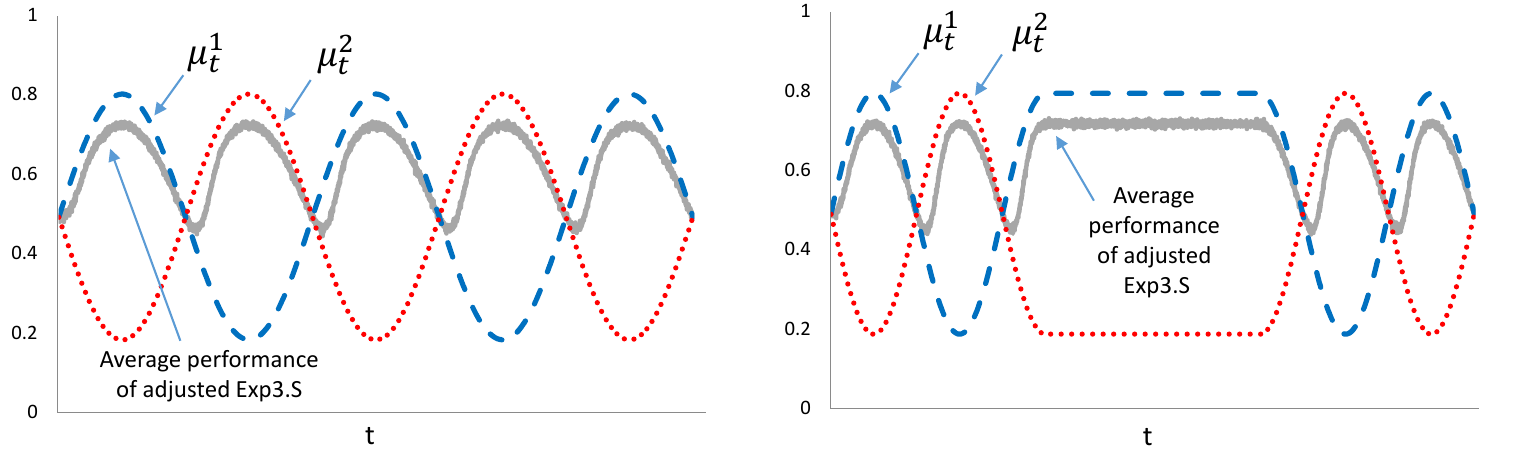}\vspace{-0.3cm}
\caption{\small Numerical simulation of the average performance trajectory of the adjusted Exp3.S policy in two complementary non-stationary mean-reward instances: ({\em Left\/}) An instances with sinusoidal expected rewards, with a fixed variation budget $V_{T}=3$. ({\em Right\/}) An instance in which similar sinusoidal evolution is limited to the first and last thirds of the horizon. In both of the instances the average performance trajectory of the policy is generated along $T=1,500,000$ epochs.
\label{fig:sim01}}\vspace{-0.2cm}
\end{figure}

\paragraph*{Experiment 1:  Fixed variation and sensitivity to time horizon.} The objective is to measure the growth rate of the regret as a function of the horizon length, under a fixed variation budget. We use two basic instances. In the first instance (displayed on the left side of Figure~\ref{fig:variation} on page \pageref{fig:variation}) the expected rewards are sinusoidal:\vspace{-0.0cm}
\begin{equation}\label{eq:sinus}
\mu^{1}_{t} = \frac{1}{2} + \frac{3}{10}\sin\left(\frac{5V_{T}\pi t}{3T}\right), \quad \quad
\mu^{2}_{t} = \frac{1}{2} + \frac{3}{10}\sin\left(\frac{5V_{T}\pi t}{3T} + \pi\right)\vspace{-0.0cm}
\end{equation}
for all $t \in\mathcal{T}$. In the second instance (depicted on the right side of Figure~\ref{fig:variation})  a similar sinusoidal evolution %of the expected reward
is limited to the first and last thirds of the horizon, where in the middle third mean rewards are constant:\vspace{-0.0cm}
\begin{displaymath}
   \mu^{1}_{t} = \left\{
     \begin{array}{lr}
       \frac{1}{2} + \frac{3}{10}\sin\left(\frac{15V_{T}\pi t}{2T}\right) & \text{ if } t < \frac{T}{3} \\
       \frac{4}{5} & \text{ if } \frac{T}{3} \leq t \leq \frac{2T}{3} \\
       \frac{1}{2} + \frac{3}{10}\sin\left(\frac{15V_{T}\pi (t-T/3)}{2T} \right) & \text{ if }  \frac{2T}{3} < t \leq T
     \end{array}
   \right.\quad\;
   \mu^{2}_{t} = \left\{
     \begin{array}{lr}
       \frac{1}{2} + \frac{3}{10}\sin\left(\frac{15V_{T}\pi t}{2T} + \pi \right) & \text{ if } t < \frac{T}{3} \\
       \frac{1}{5} & \text{ if } \frac{T}{3} \leq t \leq \frac{2T}{3} \\
       \frac{1}{2} + \frac{3}{10}\sin\left(\frac{15V_{T}\pi (t-T/3)}{2T} + \pi \right) & \text{ if } \frac{2T}{3} < t \leq T
     \end{array}
   \right.\vspace{-0.0cm}
\end{displaymath}
for all $t \in\mathcal{T}$. Both instances describe different changing environments under  the same fixed variation budget $V_{T} = 3$. In the first instance the variation budget is ``spent'' more evenly throughout the horizon, while in the second instance that budget is ``spent'' only over the first third of the horizon. For different values of $T$ up to $3\cdot 10^8$, we use 100  replications to estimate the expected regret.  The average performance trajectory (as a function of time) of the adjusted Exp3.S policy for $T=1.5\cdot 10^6$ is depicted using the solid curves in Figure~\ref{fig:sim01} (the dotted and dashed curves correspond to the mean reward paths for each arm, respectively).

The first simulation experiment depicts the average performance of the policy (as a function of time) and illustrates the  balance between exploration, exploitation, and the degree to which the policy needs to forget ``stale" information. The policy selects the arm with the highest expected reward with higher probability. Of note are the delays in identifying the cross-over points which are evidently minor, the speed at which the policy switches to the new optimal action, and the fact that the policy keeps experimenting with the sub-optimal arm. These imply that the performance  does not match the one of the dynamic oracle, but rather falls short of  the latter.

\begin{figure}[t]
\centering
\includegraphics[height=2.0in]{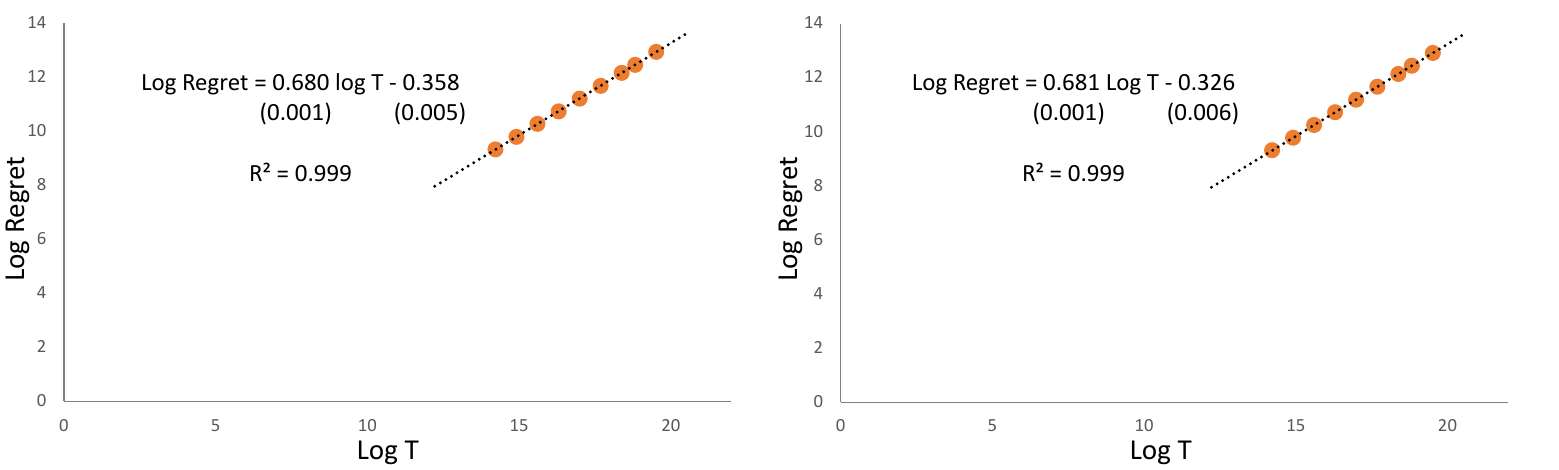}\vspace{-0.3cm}
\caption{\small Log-log plots of the averaged regret incurred by the adjusted Exp3.S as a function of the horizon length~$T$, with the resulting linear relationship (slope, and intercept) estimates under the two instances that appear in Figure \ref{fig:sim01}: ({\em Left\/}) Instance with sinusoidal expected rewards, with a fixed variation budget $V_{T}=3$. ({\em Right\/}) An instance in which similar sinusoidal evolution is limited to the first and last thirds of the horizon. Standard errors appear in parentheses.
\label{fig:sim02}}\vspace{-0.2cm}
\end{figure}

The two graphs in Figure~\ref{fig:sim02} depict  log-log plots of the mean regret as a function of the horizon. We observe that the slope is close to 2/3 and hence  consistent with the result of Theorem \ref{thm:resregret} applied to a constant variation. The standard errors for the slope and intercept estimates are in parentheses.

\paragraph*{Experiment 2:  Fixed time horizon and sensitivity to variation.} The objective of the second experiment is to measure how regret growth rate (as a function of $T$) depends on the variation budget. For this purpose we take $V_{T} = 3T^{^{\beta}}$ and we explore the dependence of the regret on $\beta$. Under the sinusoidal variation instance in (\ref{eq:sinus}), Table \ref{tb:slopes-b} reports the estimates of regret rates obtained via slopes of the log-log plots for values of $\beta$ between $0$ (constant variation, simulated in Experiment 1) and $0.5$.\footnote{We focus here on $\beta$ values up to 0.5 as for high values of $\beta$,  observing the asymptotic regret  requires a horizon significantly longer than $3\cdot 10^8$ that is used here.}

This simulation experiment illustrates how the variation level affects the policy's performance. The slopes of log-log dependencies of the regret as a function of the horizon length were estimated for the various $\beta$ values and are summarized in Table \ref{tb:slopes-b},  along with standard errors.  This is contrasted with the theoretical rates for the minimax regret obtained in previous results (Theorems \ref{thm:low} and \ref{thm:resregret}).
\begin{table}[t]
\footnotesize
 \def\arraystretch{1.25}
  \begin{center}
 %   \begin{tabular}{|C{2.5cm}|C{2.5cm}|C{2.5cm}|}
        \begin{tabular}{|c|c|c|}
    \hline
$\beta$ & Theoretical slope: &Estimated \\ % & Estimated slope: &Estimated slope:\\
 value &    $(2+\beta)/3$ &  slope \\ %Adjusted Exp3.S \\ % & Original Exp3.S & SWUCB1\\
\hline
$0.0$ & $ 0.67 $ & 0.680\;(0.001)\\% & 0.553\;(0.001)&0.655\;(0.003)\\
$0.1$ & $ 0.70 $ & 0.710\;(0.001)\\% & 0.562\;(0.001)&0.655\;(0.003)\\
$0.2$ & $ 0.73 $ & 0.730\;(0.001)\\% & 0.651\;(0.001)&0.655\;(0.003)\\
$0.3$ & $ 0.77 $ & 0.766\;(0.001)\\% & 0.783\;(0.001)&0.656\;(0.003)\\
$0.4$ & $ 0.80 $ & 0.769\;(0.001)\\% & 0.988\;(0.001)&0.749\;(0.017)\\
$0.5$ & $ 0.83 $ & 0.812\;(0.001)\\% & 1.000\;(0.001)&0.924\;(0.019)\\
%$0.6$ & $ 0.87 $ & 0.924\;(0.009) & 1.000\;(0.001)&0.972\;(0.044)\\
%$0.7$ & $ 0.90 $ & 1.000\;(0.001) & 1.000\;(0.001)&0.977\;(0.042)\\
%$0.8$ & $ 0.93 $ & 1.000\;(0.001) & 1.000\;(0.001)&0.979\;(0.037)\\
%$0.9$ & $ 0.97 $ & 1.000\;(0.001) & 1.000\;(0.001)&0.997\;(0.003)\\
\hline
    \end{tabular}%\vspace{-0.1cm}
    \caption{\small Estimated log-log slopes for growing variation budgets of the structure $V_{T}=3T^{\beta}$ (standard errors appear in parentheses),  contrasted with the slopes ($T$ dependence) for the theoretical minimax regret rates.}
    \label{tb:slopes-b}
    \end{center}\vspace{-0.35cm}
\end{table}
The estimated slopes illustrate the growth of regret as variation increases, and in that sense, Table \ref{tb:slopes-b} emphasizes the spectrum of minimax regret rates (of order $V_{T}^{1/3} T^{2/3}$) that are obtained for different variation levels. The numerical performance of the proposed policy achieves a regret rate that matches quite closely the theoretical values established in previous theorems.

\textbf{Remark.} As mentioned earlier, the Exp3.S policy was originally designed for an adversarial setting that considers a finite number of changes in the identity of the best arm, and does not have performance guarantees under the framework of the current paper, which allows for growing (horizon dependent) changes in the values of the mean rewards as well as in the identity of the best arm. In particular, when there are $H$ changes in the identity of the best arm, the upper bound that is obtained for the performance of Exp3.S in \cite{Auer-etal} is of order $H\sqrt{T}$ (excluding logarithmic terms). Notably, our experiment  illustrates a case where the number of changes of the identity of the best arm is growing with $T$. In particular, when $\beta=0.5$ the number of said changes is of order $\sqrt{T}$, thus the upper bound in \cite{Auer-etal} is of order $T$ and does not guarantee sub-linear regret anymore, in contrast with the upper bound of order $T^{5/6}$ that is established in the present paper. Additional numerical experiments implied that at settings with high variation levels the empirical performance of Exp3.S is dominated by the one achieved by the policies described in the current paper. For example, in the setting of Experiment 2, for $\beta = 0.5$ the estimated slope for Exp3.S was 1 (implying linear regret) with standard error of 0.001; similar results were obtained in other instances of high variation and frequent switches in the identity of the best arm.

\vspace{-0.0cm}
\section{Adapting to Unknown Variation}\label{sec:uppadapting}\vspace{-0.0cm}

\paragraph*{Motivation and overview.} In the previous sections we established the minimax regret rates and have put forward rate optimal policies that tune parameters using knowledge of the  variation budget $V_{T}$. This leaves open the question of designing policies that can {\it adapt} to the variation ``on the fly" without such a priori knowledge. A further issue, pertinent to this question, is the behavior of the current policy which is tuned to the ``worst case'' variation. Specifically, the proposed Rexp3 policy guarantees rate optimality by countering the worst-case sequence of mean rewards in ${\mathcal L}(V_T)$. In doing so, it ends up ``over exploring" if the realized environment is more benign. On the other hand, if the actual variation turns out to exceed the variation budget through which the policy is tuned, the policy should be expected to incur additional regret as well. To formalize these observations, one can modify the analysis of Theorem \ref{thm:resregret} to represent the regret bound in terms of both the input $V_T$ and the actual variation ${\mathcal V}(\mu;T)$, to establish that the worst case regret for Rexp3 is bounded from above as follows:\vspace{-0.1cm}
\begin{equation}\label{eq:unknown-variation}
\mathcal{R}^{\pi}(V_T,T) \;\leq\; \bar{C}(K \log K)^{1/3}T^{2/3}\cdot \max\left\{V_{T}^{1/3},\frac{{\mathcal V}(\mu;T)}{V_{T}^{2/3}}\right\},\vspace{-0.1cm}
\end{equation}
where $\bar{C}$ is the same absolute constant that appears in Theorem \ref{thm:resregret}. A similar result can be obtained for the adjusted Exp3.S policy, respectively adapting the proof of Theorem 3.

From the expression in (\ref{eq:unknown-variation}) it is possible to tease out conditions under which long-run average optimality is ensured under an inaccurate input $V_T$, which does not match the actual variation ${\mathcal V}(\mu;T)$. This also quantifies the loss associated with using an inaccurate input $V_T$ as opposed to tuning the policy using the actual variation ${\mathcal V}(\mu;T)$. On the one hand, Equation (\ref{eq:unknown-variation}) implies that whenever the actual variation does not exceed the input $V_T$, long-run average optimality is guaranteed. For example, if the policy is tuned using $V_T =T^{\alpha}$ for some $\alpha \in (0,1)$, and the variation on mean rewards is zero, which is an admissible sequence contained in ${\mathcal L}(V_T)$, then the policy will incur a regret of order $T^{(2+\alpha)/3}$. While this is long-run average optimal for any $\alpha\in(0,1)$, it also includes a regret rate ``penalty" relative to the optimal regret rate of order $T^{1/2}$ that would have been achieved if it was known a priori that there would be no variation in the mean rewards.

On the other hand, Equation (\ref{eq:unknown-variation}) also implies that long-run average optimality can be guaranteed when the actual variation \emph{exceeds} the input $V_T$, as long as $V_T$ is ``close enough" to the actual variation. In particular, when the policy is tuned by $V_T =T^{\alpha}$ for some $\alpha \in (0,1)$, it guarantees long-run average optimality as long as the actual variation ${\mathcal V}(\mu;T)$ is at most of order $T^{\alpha+\delta}$ for some $\delta < (1 - \alpha)/3$. For example, if $\alpha = 0$ and $\delta = 1/4$, Rexp3 guarantees sublinear regret of order $T^{11/12}$ (accurate tuning would have guaranteed order $T^{3/4}$).

Since there are no restrictions on the rate at which the variation budget can be spent, an interesting and challenging open problem is to delineate to what extent it is possible to design adaptive policies that do not use prior knowledge of $V_T$, yet guarantee ``good" performance. In the rest of this section we indicate a possible approach to address the above issue. Ideally, one would like the regret of adaptive policies to scale with the actual variation ${\mathcal V}(\mu;T)$, rather than with the upper bound  $V_{T}$. Our approach uses an  ``envelope" policy subordinate to which are  multiple primitive  restart EXP3-type policies. Each of the latter are tuned to a different ``guess'' of the variation budget, and the ``master'' policy switches among the subordinates based on realized rewards. While we have no proof of optimality, or strong theoretical indication to believe an optimal adaptive policy is to be found in this family, we believe that the main ideas presented in our approach may be useful for deriving fully adaptive rate-optimal policies.
%While we have no proof of optimality, or strong theoretical indication to believe an optimal adaptive policy is to be found in this family, numerical results indicate that such a conjecture is plausible.

%\vspace{-0.1cm}
\subsection{An envelope policy}\label{subsec:envelope}%\vspace{-0.1cm}

We introduce an envelope policy $\bar{\pi}$ that judiciously applies $M$ subordinate MAB policies $\left\{\pi^{m}\;:\; m=1,\ldots,M\right\}$. At each epoch only one of these ``virtual" policies may be used, but the collected information can be shared among all subordinate policies (concrete subordinate MAB policies will be suggested in \S\ref{sec:poldesign}).

\vspace{-0.6cm}\begin{center}
\line(1,0){490}
\end{center}
\vspace{-6mm}\textbf{Envelope policy ($\mathbf{\bar{\pi}}$).} Inputs: $M$ admissible policies $\left\{\pi^{m}\right\}_{m=1}^{M}$, and a number $\bar{\gamma}\in\left(0,1\right]$. \vspace{-0.3cm}
\begin{enumerate}
    \item Initialization: for $t=1$, for any $m\in \left\{1,\ldots,M\right\}$ set $\nu^{m}_{t} = 1$ \vspace{-0.2cm}
    \item For each $t = 1,2,\ldots$ do\vspace{-0.2cm}
  \begin{itemize}
  \item For each $m\in\left\{1,\ldots,M\right\}$, set\vspace{-0.4cm}
    \[
    q^{m}_{t} \;=\; \left(1-\bar{\gamma}\right)\frac{\nu^{m}_{t}}{\sum_{m'=1}^{M}\nu^{m'}_{t}} + \frac{\bar{\gamma}}{M}\vspace{-0.3cm}
    \]
    \item Draw $m'$ from $\left\{1,\ldots,M\right\}$ according to the distribution $\left\{q^{m}_{t}\right\}_{m=1}^{M}$\vspace{-0.1cm}
    \item Select the arm $\hat{k} = \pi^{m'}_t$ from the set $\left\{1,\ldots,K\right\}$ and receive a reward $X^{\hat{k}}_{t}$
        \vspace{-0.1cm}
    \item Set $\hat{Y}^{m'}_{t} = X^{\hat{k}}_{t}/q^{m'}_{t}$, and for any $m\neq m'$ set $\hat{Y}^{m}_{t} = 0$ \vspace{-0.1cm}
    \item For all $m\in\left\{1,\ldots,M\right\}$ update:\vspace{-5mm}
        \[
\nu^{m}_{t+1}\; =\; \nu^{m}_{t}\exp\left\{ \frac{\bar{\gamma} \hat{Y}^{m}_{t}}{M}  \right\}.
  \vspace{-4mm}
  \]
  \item Update subordinate policies $\left\{\pi^{m}\;:\; m=1,\ldots,M\right\}$; details of subordinate policy structure appear below.
  \end{itemize}
\end{enumerate}
\begin{center}\vspace{-7mm}
\line(1,0){490}
\end{center}\vspace{-4mm}

The envelope policy operates as follows. At each epoch a subordinate policy is drawn from a distribution $\left\{q_{t}^{m}\right\}_{m=1}^{M}$ that is updated  every epoch based on the observed rewards. This distribution is endowed  with an exploration rate $\bar{\gamma}$ that is used to balance exploration and exploitation over the subordinate policies.

\vspace{-0.1cm}
\subsection{Structure of the subordinate policies}\label{sec:poldesign}\vspace{-0.1cm}

In \S\ref{sec:upp2} we already identified classes of candidate Rexp3 and adjusted Exp3.S policies, that were shown to achieve rate optimality when \emph{accurately tuned} ex ante, using the bound $V_{T}$.  We therefore take  $(\pi^{1},\ldots,\pi^{m})$, the subordinate policies,  to be the proposed  adjusted Exp3.S policies tuned by exploration rates $(\gamma_{1}\ldots,\gamma_{M})$ to be specified. We denote by $\bar{\pi}_{t}\in\left\{1,\ldots, M\right\}$ the action of the envelope policy at time $t$, and by $\pi^{m}_{t}\in\left\{1,\ldots,K\right\}$ the action of policy $\pi^{m}$ at the same epoch. We denote by $\left\{p_{t}^{k,m}\;:\; k=1,\ldots,K\right\}$ the distribution from which $\pi^{m}_{t}$ is drawn, and by $\left\{w_{t}^{k,m}\;:\; k=1,\ldots,K\right\}$ the weights associated with this distribution according to the Rexp3.S  structure. We adjust the Rexp3.S description that is given in \S\ref{sec:upp3} by defining the following update rule. At each epoch $t$, and for each arm $k$, each subordinate Rexp3.S policy $\pi^{m}$ selects an arm in $\left\{1,\ldots,K\right\}$ according to the distribution $\left\{p_{t}^{k,m}\right\}$. However, rather than updating the weights $\left\{w_{t}^{k,m}\right\}$ using the observation from that arm, each subordinate policy uses the update rule:\vspace{-0.0cm}
\[
\hat{X}^{k}_{t} \;=\; \sum_{m=1}^{M}\frac{X^{k}_{t}}{p^{k,m}_{t}} \mathbbm{1}\left\{\bar{\pi}_{t}=m \right\}\mathbbm{1}\left\{\pi^{m}_{t}=k \right\},\quad k=1,\ldots,K\vspace{-0.0cm}
\]
and then updates weights as in the original Rexp3.S description:\vspace{-0mm}
\[
w^{k,m}_{t+1} \;=\; w^{k,m}_{t}\exp\left\{ \frac{\gamma_{m} \hat{X}^{k}_{t}}{K} \right\} +  \frac{e\alpha}{K}\sum_{k'=1}^{K}w^{k',m}_{t}.
\]
We note that at each time step the variables $\hat{X}^{k}_{t}$ are updated in an identical manner by all the subordinate Rexp3.S policies (independent of $m$). This update rule implies that information is shared between subordinate  policies.

\subsection{Numerical Analysis}

We consider a case where there are three possible levels for the realized variation ${\mathcal V}(\mu;T)$. With slight abuse of notation, we write $ {\mathcal V}(\mu;T) \in \{V_{T,1},V_{T,2},V_{T,3}\}$,  with $V_{T,1}=0$ (no variation), $V_{T,2}=3$ (fixed variation), and $V_{T,3}=3T^{0.2}$ (increasing variation). The envelope policy does not know which of the latter variation levels describes the actual variation. We measured the performance of the envelope policy, tuned by $\bar{\gamma} = \min\left\{1, \sqrt{M\log\left(M\right)\cdot(e-1)^{-1}\cdot T^{-1}}\right\}$, and with three input policies that ``guess" $V_{T,1}=0$, $V_{T,2}=3$, and $V_{T,3}=3T^{0.2}$, respectively. More formally, the input policies are selected as follows: $\pi_{1}$ being the Exp3.S policy with $\alpha= 1/T$ and $\gamma_{1} = \min\left\{1, \sqrt{K\log\left(KT\right)\cdot T^{-1}}\right\}$ (the parametric values that appear in \citealt{Auer-etal}); $\pi_{2}$ and $\pi_{3}$ are both Exp3.S policies with $\alpha = 1/T$ and $\gamma_{m} = \min\left\{1, \left(\frac{2V_{T,m}K\log\left(KT\right)}{(e-1)^{2}T}\right)^{1/3}\right\}$ with the respective $V_{T,m}$ values. Each subordinate Exp3.S policy uses the update rule described in the previous paragraph.

We measure the performance of the envelope policy in three different settings, corresponding to the different values of the variation ${\mathcal V}(\mu;T)$ specified above. When the realized variation is $\mathcal{V}(\mu;T) = V_{T,1} = 0$ (no variation), the mean rewards were  $\mu_{t}^{1} = 0.2$ and $\mu_{t}^{2} = 0.8$ for all $t= 1,\ldots, T$. When the variation is $\mathcal{V}(\mu;T) = V_{T,2} = 3$ (fixed variation), the mean rewards followed the sinusoidal variation instance in (\ref{eq:sinus}). When the variation is $\mathcal{V}(\mu;T) = V_{T,3} = 3T^{0.2}$ (increasing variation), the mean rewards followed the same sinusoidal pattern where variation increased with the horizon length. We then repeat the experiment while considering an envelope policy with only two subordinate policies, one that ``guesses" $V_{T,1}=0$ (no variation) and one that guesses $V_{T,3}=3T^{0.2}$ (increasing variation), but measuring its performance under each of the cases.

\begin{figure}[h!]
\centering
\includegraphics[height=2.1in]{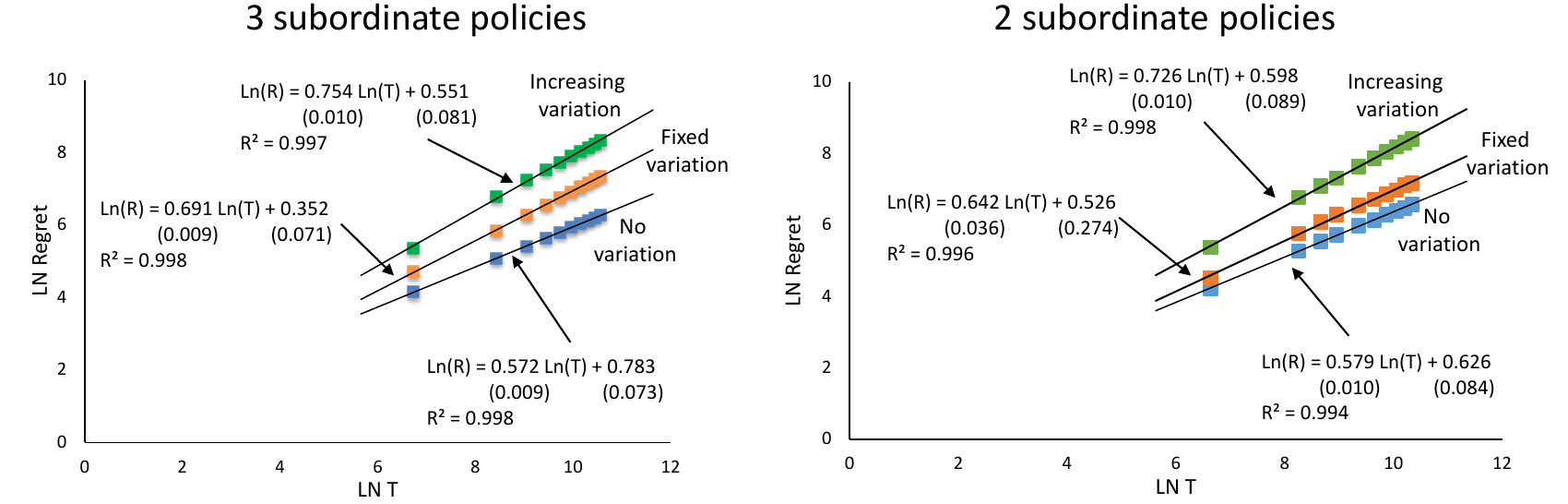}\vspace{-0.2cm}
\caption{\small
Adapting to unknown variation: Log-log plots of regret as a function of the horizon length $T$ obtained under three different variation levels: no variation, fixed variation, and increasing variation of the form $V_T = 3T^{0.2}$. \emph{(Left)} Performance of an envelope policy with three subordinate policies, each corresponding to one of the former variation levels. \emph{(Right)} Performance of the envelope policy with two subordinate policies: one corresponding to no variation and once corresponding to the increasing variation case (standard errors appear in parentheses).
\label{fig:sim3}}\vspace{-0.2cm}
\end{figure}

\textbf{Discussion.} The left side of Figure~\ref{fig:sim3} depicts the three log-log plots obtained for the three scenarios. The slopes illustrates that the envelope policy  seems to adapt to different realized variation levels. In particular, the slopes appear close to those that  would have been obtained with prior knowledge of the variation level. However, the uncertainty regarding the realized variation may cause a larger multiplicative constant in the regret expression; this is demonstrated by the higher intercept in the log-log plot of the fixed variation case (0.352) relative to the intercept obtained when the same variation level is a priori known (-0.358); see  left part of Figure~\ref{fig:sim02}.

The envelope policy seems effective when the set of possible variations is known a priori. The second part of the experiment suggests that a similar approach may be effective even when one cannot limit a priori the set of realized variations. The  right side of Figure~\ref{fig:sim3} depicts the three log-log plots obtained under the three cases when the envelope policy uses only two subordinate policies: one that ``guesses" no variation, and one that ``guesses" an increasing variation. While performance was sensitive to the initial amplitude of the sinusoidal functions, the results suggest that, on average, the envelope policy achieved comparable performance to the one using three subordinate policies, which include the well-specified fixed variation case.

\appendix

\vspace{-0.0cm}
\section{Proofs of main results}\label{app:proofs}
\vspace{-0.0cm}
\textbf{Proof of Theorem \ref{thm:low}.}
At a high level the proof adapts a general approach of identifying a worst-case nature ``strategy" (see proof of Theorem 5.1 in \cite{Auer-etal}, which analyze the worst-case regret relative to a single best action benchmark in a fully adversarial environment), extending these ideas appropriately to our setting. Fix $T\geq 1$, $K\geq 2$, and $V_{T}\in \left[K^{-1},K^{-1}T\right]$. In what follows we restrict nature to the class $\mathcal{V}'\subseteq \mathcal{V}$ that was described in \S3, and show that when $\mu$ is drawn randomly from $\mathcal{V}'$, any policy in $\mathcal{P}$ must incur regret of order $\left(KV_{T}\right)^{1/3}T^{2/3}$.

\textbf{Step 1 (Preliminaries).} Define a partition of the decision horizon $\mathcal{T}$ to $m = \left\lceil \frac{T}{\Delta} \right\rceil$ batches $\mathcal{T}_{1},\ldots,\mathcal{T}_{m}$ of size $\Delta$ each (except perhaps $\mathcal{T}_{m}$) according to $\mathcal{T}_{j} \;=\; \left\{ t\; :\; (j-1)\Delta + 1 \leq t \leq \min \left\{ j\Delta,T\right\} \right\}$ for all $j = 1,\ldots,m,$ where $m = \lceil T/\Delta \rceil$ is the number of batches. For some $\varepsilon > 0$ that will be specified shortly, define $\mathcal{V}'$ to be the set of reward vectors sequences $\mu$ such that:\vspace{-0.2cm}
\begin{itemize}
  \item $\mu_{t}^{k}\in \left\{1/2,1/2+\varepsilon\right\}$ for all $k\in\mathcal{K}$, $t\in \mathcal{T}$\vspace{-0.2cm}
  \item $\sum_{k\in \mathcal{K}}\mu_{t}^{k} = K/2 + \varepsilon$ for all $t\in \mathcal{T}$\vspace{-0.2cm}
  \item $\mu_{t}^{k} = \mu_{t+1}^{k}$ for any $(j-1)\Delta+1\leq t \leq \min\left\{j\Delta,T\right\}-1$, $j=1,\ldots,m$, for all $k\in \mathcal{K}$\vspace{-0.2cm}
\end{itemize}
For each sequence in $\mathcal{V}'$ in any epoch there is exactly one arm with expected reward $1/2+\varepsilon$ where the rest of the arms have expected reward $1/2$, and expected rewards cannot change within a batch.
Let $\varepsilon = \min\left\{\frac{1}{4}\cdot\sqrt{K/\Delta}, V_{T}\Delta/T\right\}$. Then, for any $\mu\in \mathcal{V}'$ one has:\vspace{-0.2cm}
\[
\sum_{t=1}^{T-1}\sup_{k\in \mathcal{K}}\left|\mu_{t}^{k} - \mu_{t+1}^{k}\right|
\;\leq\; \sum_{j=1}^{m-1}\varepsilon
\;=\;\left(\left\lceil \frac{T}{\Delta} \right\rceil - 1\right)\cdot \varepsilon
\;\leq \frac{T\varepsilon}{\Delta}_{T}
\;\leq \; V_{T},\vspace{-0.2cm}
\]
where the first inequality follows from the structure of $\mathcal{V}'$. Therefore, $\mathcal{V}'\subset \mathcal{V}$.

\textbf{Step 2 (Single batch analysis).} Fix some policy $\pi \in \mathcal{P}$, and fix a batch $j\in\left\{1,\ldots,m\right\}$. Let $k_{j}$ denote the ``good" arm of batch $j$. We denote by $\mathbb{P}^{j}_{k_{j}}$ the probability distribution conditioned on arm $k_{j}$ being the ``good" arm in batch $j$, and by $\mathbb{P}_{0}$ the probability distribution with respect to random rewards (i.e. expected reward $1/2$) for each arm. We further denote by $\mathbb{E}^{j}_{k_{j}}[\cdot]$ and $\mathbb{E}_{0}[\cdot]$ the respective expectations. Assuming binary rewards, we let $X$ denote a vector of $\left|\mathcal{T}_{j}\right|$ rewards, i.e. $X\in \left\{0,1\right\}^{\left|\mathcal{T}_{j}\right|}$. We denote by $N^{j}_k$ the number of times arm $k$ was selected in batch $j$. In the proof we use Lemma A.1 from \cite{Auer-etal} that characterizes the difference between the two different expectations of some function of the observed rewards vector:

\begin{lemma}\label{lem:boundf}\textbf{\textup{(Lemma A.1 from \citealt{Auer-etal})}}
Let $f:\left\{0,1\right\}^{\left|\mathcal{T}_{j}\right|}\rightarrow \left[0,M\right]$ be a bounded real function. Then, for any $k\in\mathcal{K}$:\vspace{-0.2cm}
\[
\mathbb{E}^{j}_{k}\left[f(X)\right] - \mathbb{E}_{0}\left[f(X)\right]
 \leq \frac{M}{2}\sqrt{- \mathbb{E}_{0}\left[N^{j}_{k}\right]\log\left(1 - 4\varepsilon^{2}\right)}.
\]
\end{lemma}
Recalling that $k_{j}$ denotes the ``good" arm of batch $j$, one has\vspace{-0.1cm}
\[
\mathbb{E}^{j}_{k_{j}}\left[\mu_{t}^{\pi}\right]
 \;=\; \left(\frac{1}{2} + \varepsilon \right)\mathbb{P}^{j}_{k_{j}}\left\{\pi_t = k_{j}\right\} + \frac{1}{2}\mathbb{P}^{j}_{k_{j}}\left\{\pi_t \neq k_{j}\right\}
 \;=\; \frac{1}{2} + \varepsilon \mathbb{P}^{j}_{k_{j}}\left\{\pi_t = k_{j}\right\},\vspace{-0.1cm}
\]
and therefore,
\begin{equation}\label{eq:klow1}
\mathbb{E}^{j}_{k_{j}}\left[\sum_{t\in\mathcal{T}_{j}}\mu_{t}^{\pi}\right]
 \;=\; \frac{\left|\mathcal{T}_{j}\right|}{2} + \sum_{t\in\mathcal{T}_{j}}\varepsilon \mathbb{P}^{j}_{k_{j}}\left\{\pi_t = k_{j}\right\}
 \;=\; \frac{\left|\mathcal{T}_{j}\right|}{2} + \varepsilon\mathbb{E}^{j}_{k_{j}}\left[N^{j}_{k_{j}}\right].
\end{equation}
In addition, applying Lemma \ref{lem:boundf} with $f(X) = N^{j}_{k_{j}}$ (clearly $N^{j}_{k_{j}} \in \left\{0,\ldots,\left|\mathcal{T}_{j}\right|\right\}$) we have:\vspace{-0.1cm}
\[
\mathbb{E}^{j}_{k_{j}}\left[N^{j}_{k_{j}}\right]
 \;\leq\; \mathbb{E}_{0}\left[N^{j}_{k_{j}}\right] + \frac{\left|\mathcal{T}_{j}\right|}{2}\sqrt{- \mathbb{E}_{0}\left[N^{j}_{k_{j}}\right]\log\left(1 - 4\varepsilon^{2}\right)}.\vspace{-0.1cm}
\]
Summing over arms, one has:\vspace{-0.2cm}
\begin{eqnarray}
\sum_{k_{j}=1}^{K}\mathbb{E}^{j}_{k_{j}}\left[N^{j}_{k_{j}}\right]
 &\leq& \sum_{k_{j}=1}^{K} \mathbb{E}_{0}\left[N^{j}_{k_{j}}\right]
 + \sum_{k_{j}=1}^{K} \frac{\left|\mathcal{T}_{j}\right|}{2}\sqrt{- \mathbb{E}_{0}\left[N^{j}_{k_{j}}\right]\log\left(1 - 4\varepsilon^{2}\right) }\notag\\
&\leq& \left|\mathcal{T}_{j}\right| + \frac{\left|\mathcal{T}_{j}\right|}{2}\sqrt{ - \log\left(1 - 4\varepsilon^{2}\right)\left|\mathcal{T}_{j}\right| K  },%\notag\\
%&\stackrel{(b)} \leq& \Delta + \frac{\Delta}{2}\sqrt{ - \log\left(1 - 4\varepsilon^{2}\right)\Delta K  },
\label{eq:klow1.1}
\end{eqnarray}
for any $j\in \left\{1,\ldots,m\right\}$, where the last inequality holds since $\sum_{k_{j}=1}^{K} \mathbb{E}_{0}\left[N^{j}_{k_{j}}\right] = \left|\mathcal{T}_{j}\right|$, and thus by Cauchy-Schwarz inequality $\sum_{k_{j}=1}^{K} \sqrt{\mathbb{E}_{0}\left[N^{j}_{k_{j}}\right]} \leq~\sqrt{\left|\mathcal{T}_{j}\right| K}$.

\textbf{Step 3 (Regret along the horizon).}
Let $\tilde{\mu}$ be a random sequence of expected rewards vectors, in which in every batch the ``good" arm is drawn according to an independent uniform distribution over the set $\mathcal{K}$. Clearly, every realization of $\tilde{\mu}$ is in $\mathcal{V}'$. In particular, taking expectation over $\tilde{\mu}$, one has:\vspace{-0.2cm}
\begin{eqnarray}
\mathcal{R}^{\pi}(\mathcal{V}',T)
 &=& \sup_{\mu\in\mathcal{V}'}\left\{\sum_{t=1}^{T}\mu^{\ast}_t
 - \mathbb{E}^{\pi}\left[\sum_{t=1}^{T}\mu^{\pi}_t\right]\right\}
 \;\geq\; \mathbb{E}^{\tilde{\mu}}\left[\sum_{t=1}^{T}\tilde{\mu}^{\ast}_t
 - \mathbb{E}^{\pi}\left[\sum_{t=1}^{T}\tilde{\mu}^{\pi}_t\right]\right]\notag\\
&\geq& \sum_{j=1}^{m}\left(\sum_{t\in\mathcal{T}_{j}}\left(\frac{1}{2} + \varepsilon\right)
 - \frac{1}{K}\sum_{k_{j}=1}^{K}\mathbb{E}^{\pi}\mathbb{E}^{j}_{k_{j}}\left[\sum_{t\in\mathcal{T}_{j}}\tilde{\mu}_{t}^{\pi}\right]\right)\notag\\
 &\stackrel{(a)} \geq& \sum_{j=1}^{m}\left(\sum_{t\in\mathcal{T}_{j}}\left(\frac{1}{2} + \varepsilon\right)
  - \frac{1}{K}\sum_{k_{j}=1}^{K}\left(\frac{\left|\mathcal{T}_{j}\right|}{2} + \varepsilon\mathbb{E}^{\pi}\mathbb{E}^{j}_{k_{j}}\left[N^{j}_{k_{j}}\right]\right)\right)\notag\\
 &\geq& \sum_{j=1}^{m}\left(\sum_{t\in\mathcal{T}_{j}}\left(\frac{1}{2} + \varepsilon\right)
 - \frac{\left|\mathcal{T}_{j}\right|}{2}
  - \frac{\varepsilon}{K}\mathbb{E}^{\pi}\sum_{k_{j}=1}^{K}\mathbb{E}^{j}_{k_{j}}\left[N^{j}_{k_{j}}\right]\right)\notag\\
%&\stackrel{(b)} \geq& \sum_{j=1}^{m}\left(\left|\mathcal{T}_{j}\right|\varepsilon
%  - \frac{\varepsilon}{K}\left(  \Delta + \frac{\Delta}{2}\sqrt{- \log\left(1 - %4\varepsilon^{2}\right)\Delta K  } \right)\right)\notag\\
&\stackrel{(b)}\geq &  T\varepsilon
 - \frac{T \varepsilon}{K}
 - \frac{T\varepsilon}{2K}\sqrt{- \log\left(1 - 4\varepsilon^{2}\right)\Delta K  }\notag\\
 &\stackrel{(c)}\geq &\frac{T\varepsilon}{2}
 - \frac{T\varepsilon^{2}}{K}\sqrt{\log\left(4/3\right)\Delta K  },\notag
\end{eqnarray}
where: $(a)$ holds by (\ref{eq:klow1}); $(b)$ holds by (\ref{eq:klow1.1}), since $\sum_{j=1}^{m}\left|\mathcal{T}_{j}\right| = T$, since $m \geq T/\Delta$, and since $\left|\mathcal{T}_{j}\right| \leq \Delta$ for all $j\in\left\{1,\ldots, m\right\}$; and $(c)$ holds by $4\varepsilon^{2} \leq 1/4$, and $-\log(1-x)\leq 4\log(4/3)x$ for all $x\in\left[0,1/4\right]$, and since $K\geq 2$.
Set $\Delta =~\left \lceil K^{1/3}\left(\frac{T}{V_{T}}\right)^{2/3}\right \rceil$. Since $\varepsilon = \min\left\{\frac{1}{4}\cdot\sqrt{K/\Delta}_{T}, V_{T}\Delta/T\right\}$, one has:\vspace{-0.2cm}
\begin{eqnarray}
\mathcal{R}^{\pi}(\mathcal{V}',T)
   &\geq& T\varepsilon\left(\frac{1}{2} - \varepsilon\sqrt{\frac{\Delta_T \log(4/3)}{K}}   \right)
   \;\geq\; T \varepsilon\left(\frac{1}{2} - \frac{\sqrt{\log(4/3)}}{4}\right)\notag\\
   &\geq& \frac{1}{4}\cdot \min\left\{\frac{T}{4}\cdot\sqrt{\frac{K}{\Delta}}
   , V_{T}\Delta\right\}\notag\\
   &\geq& \frac{1}{4}\cdot \min\left\{\frac{T}{4}\cdot\sqrt{\frac{K}{2K^{1/3}(T/V_{T})^{2/3}}}
   , \left(KV_{T}\right)^{1/3}T^{2/3}\right\}\notag\\
  &\geq& \frac{1}{4\sqrt{2}}\cdot(K V_T)^{1/3}  T^{2/3}.\notag
\end{eqnarray}
This concludes the proof.\qed

\paragraph*{Proof of Theorem \ref{thm:resregret}.}
The structure of the proof is as follows. First, we break the horizon to a sequence of batches of size $\Delta$ each, and analyze the performance gap between the single best action and the dynamic oracle in each batch. Then, we plug in a known performance guarantee for Exp3 relative to the single best action, and sum over batches to establish the regret of Rexp3 relative to the dynamic oracle.

\textbf{Step 1 (Preliminaries).}
Fix $T\geq 1$, $K\geq 2$, and $V_{T}\in \left[K^{-1}, K^{-1}T\right]$. Let $\pi$ be the Rexp3 policy, tuned by $\gamma = \min\left\{1 \;,\; \sqrt{\frac{K\log K}{(e-1)\Delta}} \right\}$ and $\Delta\in \left\{1,\ldots,T\right\}$ (to be specified later on). Define a partition of the decision horizon $\mathcal{T}$ to $m = \left\lceil \frac{T}{\Delta} \right\rceil$ batches $\mathcal{T}_{1},\ldots,\mathcal{T}_{m}$ of size $\Delta$ each (except perhaps $\mathcal{T}_{m}$) according to $\mathcal{T}_{j} \;=\; \left\{ t\; :\; (j-1)\Delta + 1 \leq t \leq \min \left\{ j\Delta,T\right\} \right\}$ for all $j = 1,\ldots,m,$ where $m = \lceil T/\Delta \rceil$ is the number of batches. Let $\mu \in \mathcal{V}$, and fix $j\in\left\{1,\ldots,m\right\}$. We decomposition the regret in batch $j$:\vspace{-0.1cm}
\begin{equation}\label{eq:decomposeres}\small
\mathbb{E}^{\pi}\left[\sum_{t \in \mathcal{T}_{j}}\left(\mu^{\ast}_{t} - \mu^{\pi}_{t}\right)\right]
 \;=\; \underbrace{\sum_{t \in \mathcal{T}_{j}}\mu^{\ast}_{t}
 - \mathbb{E}\left[\max_{k\in \mathcal{K}}\left\{\sum_{t \in \mathcal{T}_{j}}X^{k}_{t}\right\}\right]}_{J_{1,j}}
 + \underbrace{\mathbb{E}\left[\max_{k\in \mathcal{K}}\left\{\sum_{t \in \mathcal{T}_{j}}X^{k}_{t}\right\}\right]
  -  \mathbb{E}^{\pi}\left[\sum_{t \in \mathcal{T}_{j}}\mu^{\pi}_{t}\right]}_{J_{2,j}}.\vspace{-0.1cm}
\end{equation}
The first component, $J_{1,j}$, is the expected loss associated with using a single action over batch $j$. The second component, $J_{2,j}$, is the expected regret relative to the best static action in batch $j$.

\textbf{Step 2 (Analysis of $J_{1,j}$ and $J_{2,j}$).}
Defining $\mu_{T+1}^{k} = \mu_{T}^{k}$ for all $k\in \mathcal{K}$, we denote the variation in expected rewards along batch $\mathcal{T}_{j}$ by $V_{j} = \sum_{t \in \mathcal{T}_{j}}\max_{k\in \mathcal{K}}\left|\mu^{k}_{t+1} - \mu^{k}_{t}\right|$. We note that:\vspace{-0.2cm}
\begin{equation}\label{eq:variation budgres}
\sum_{j=1}^{m}V_{j}
 \;=\; \sum_{j=1}^{m}\sum_{t \in \mathcal{T}_{j}}\max_{k\in \mathcal{K}}\left|\mu^{k}_{t+1} - \mu^{k}_{t}\right|
 \;\leq\; V_{T}.\vspace{-0.2cm}
\end{equation}
Let $k_{0}$ be an arm with best expected performance over $\mathcal{T}_j$: $k_0 \in\argmax_{k\in \mathcal{K}}\left\{\sum_{t \in \mathcal{T}_{j}}\mu^{k}_{t}\right\}$. Then,\vspace{-0.2cm}
\begin{equation}\label{eq:oracles}
\max_{k\in \mathcal{K}}\left\{\sum_{t \in \mathcal{T}_{j}}\mu^{k}_{t}\right\}
 \;=\; \sum_{t \in \mathcal{T}_{j}}\mu^{k_{0}}_{t}
 \;=\; \mathbb{E}\left[\sum_{t \in \mathcal{T}_{j}}X^{k_{0}}_{t} \right]
 \;\leq\; \mathbb{E}\left[\max_{k\in\mathcal{K}}\left\{\sum_{t \in \mathcal{T}_{j}}X^{k}_{t} \right\}\right],\vspace{-0.2cm}
\end{equation}
and therefore, one has:\vspace{-0.5cm}
\begin{eqnarray}
J_{1,j}
&=& \sum_{t \in \mathcal{T}_{j}}\mu^{\ast}_{t}
 - \mathbb{E}\left[\max_{k\in \mathcal{K}}\left\{\sum_{t \in \mathcal{T}_{j}}X^{k}_{t}\right\}\right]
\;\stackrel{(a)} \leq\; \sum_{t \in \mathcal{T}_{j}}\left(\mu^{\ast}_{t} - \mu^{k_0}_{t}\right)\notag\\
&\leq& \Delta\max_{t\in \mathcal{T}_{j}} \left\{ \mu^{\ast}_{t} - \mu^{k_0}_{t}  \right\}
\;\stackrel{(b)} \leq\; 2V_{j}\Delta,\label{eq:upp4res}
\end{eqnarray}
for any $\mu\in\mathcal{V}$ and $j\in \left\{1,\ldots,m\right\}$, where $(a)$ holds by (\ref{eq:oracles}) and $(b)$ holds by the following argument: otherwise there is an epoch $t_0\in\mathcal{T}_{j}$ for which $\mu^{\ast}_{t_0} - \mu^{k_0}_{t_0} > 2V_{j}$. Indeed, suppose that $k_{1} = \argmax_{k\in \mathcal{K}}\mu_{t_{0}}^{k}$. In such case, for all $t\in \mathcal{T}_{j}$ one has $\mu^{k_{1}}_{t} \geq \mu^{k_{1}}_{t_{0}} - V_{j} > \mu^{k_0}_{t_0} + V_{j} \geq \mu^{k_0}_{t}$, since $V_{j}$ is the maximal variation in batch $\mathcal{T}_{j}$. This however, contradicts the optimality of $k_0$ at epoch $t$, and thus (\ref{eq:upp4res}) holds. In addition, Corollary 3.2 in \cite{Auer-etal} points out that the regret incurred by Exp3 (tuned by $\gamma = \min\left\{1 \;,\; \sqrt{\frac{K\log K}{(e-1)\Delta}} \right\}$) along $\Delta$ epochs, relative to the single best action, is bounded by $2\sqrt{e-1}\sqrt{\Delta K\log K}$. Therefore, for each $j\in \left\{1,\ldots,m\right\}$ one has:\vspace{-0.2cm}
\begin{equation}
J_{2,j}
 \;=\; \mathbb{E}\left[\max_{k\in \mathcal{K}}\left\{\sum_{t \in \mathcal{T}_{j}}X^{k}_{t}\right\}
 - \mathbb{E}^{\pi}\left[\sum_{t \in \mathcal{T}_{j}}\mu^{\pi}_{t}\right]\right]
\;\stackrel{(a)} \leq\; 2\sqrt{e-1}\sqrt{\Delta K\log K},\label{eq:upp2b}\vspace{-0.2cm}
\end{equation}
for any $\mu\in\mathcal{V}$, where $(a)$ holds since within each batch arms are pulled according to Exp3($\gamma$).

\textbf{Step 3 (Regret along the horizon).}
Summing over $m = \left\lceil T/\Delta\right\rceil$ batches we have:\vspace{-0.3cm}
\begin{eqnarray}\label{eq:fordisc}
\mathcal{R}^{\pi}(\mathcal{V},T)
&=& \sup_{\mu\in\mathcal{V}}\left\{\sum_{t=1}^{T}\mu^{\ast}_t - \mathbb{E}^{\pi}\left[\sum_{t=1}^{T}\mu^{\pi}_t\right]\right\}%\notag\\
\;\stackrel{(a)}\leq\; \sum_{j= 1}^{m}\left(2\sqrt{e-1}\sqrt{\Delta K\log K} + 2V_{j}\Delta\right)\notag\\
&\stackrel{(b)}\leq& \left(\frac{T}{\Delta}+1\right)\cdot 2\sqrt{e-1}\sqrt{\Delta K\log K} + 2\Delta V_{T}.\notag\\
&=& \frac{2\sqrt{e-1}\sqrt{K\log K}\cdot T}{\sqrt{\Delta}} + 2\sqrt{e-1}\sqrt{\Delta K\log K} + 2\Delta V_{T},
\end{eqnarray}
where: (a) holds by (\ref{eq:decomposeres}), (\ref{eq:upp4res}), and (\ref{eq:upp2b}); and (b) follows from (\ref{eq:variation budgres}). Selecting $\Delta =~\left\lceil \left(K\log K\right)^{1/3}\left(T/V_{T}\right)^{2/3}\right\rceil$, one has:\vspace{-0.5cm}
\begin{eqnarray}
\mathcal{R}^{\pi}(\mathcal{V},T)
&\leq& 2\sqrt{e-1}\left(K\log K \cdot V_{T}\right)^{1/3}T^{2/3}\notag\\
 & &\quad + 2\sqrt{e-1}\sqrt{\left(\left(K\log K\right)^{1/3}\left(T/V_{T}\right)^{2/3}+1\right) K\log K}\notag\\
 & &\quad + 2\left(\left(K\log K\right)^{1/3}\left(T/V_{T}\right)^{2/3}+1\right)V_{T}\notag\\
&\stackrel{(a)}\leq&  \left(\left(2+2\sqrt{2}\right)\sqrt{e-1}+4\right)\left(K\log K \cdot V_{T}\right)^{1/3}T^{2/3},\notag
\end{eqnarray}
where (a) follows from $T\geq K\geq 2$, and $V_{T}\in\left[K^{-1},K^{-1}T\right]$. This concludes the proof.\qed

\paragraph*{Proof of Theorem \ref{thm:contregret}} We prove that by selecting the tuning parameters to be $\alpha = \frac{1}{T}$ and $\gamma = \min\left\{1, \left(\frac{4V_{T}K\log\left(KT\right)}{(e-1)^{2}T}\right)^{1/3}\right\}$, Exp3.S achieves near optimal performance in the non-stationary stochastic setting. The structure of the proof is as follows: First, we break the decision horizon to a sequence of decision batches and analyze the difference in performance between the sequence of single best actions and the performance of the dynamic oracle. Then, we analyze the regret of Exp3.S relative to a sequence composed of the single best actions of each batch (this part of the proof roughly follows the proof lines of Theorem $8.1$ in \cite{Auer-etal}, while considering a possibly infinite number of changes in the identity of the best arm). Finally, we select tuning parameters that minimize the overall regret.

\textbf{Step 1 (Preliminaries).} Fix $T\geq 1$, $K\geq 2$, and $V_{T}\in \left[K^{-1}, K^{-1}T\right]$. Let $\pi$ be the Exp3.S policy described in \S\ref{sec:upp2}, tuned by $\alpha = \frac{1}{T}$ and $\gamma = \min\left\{1, \left(\frac{4V_{T}K\log\left(KT\right)}{(e-1)^{2}T}\right)^{1/3}\right\}$, and let $\Delta\in \left\{1,\ldots,T\right\}$ be a batch size (to be specified later on). We break the horizon~$\mathcal{T}$ into a sequence of batches $\mathcal{T}_{1},\ldots,\mathcal{T}_{m}$ of size $\Delta$ each (except, possibly $\mathcal{T}_{m}$) according to: $\mathcal{T}_{j} \;=\; \left\{ t\; :\; (j-1)\Delta + 1 \leq t \leq \min \left\{ j\Delta,T\right\} \right\}$, $j = 1,\ldots,m$. Let $\mu \in \mathcal{V}$, and fix $j\in\left\{1,\ldots,m\right\}$. We decompose the regret in batch $j$:\vspace{-0.2cm}
\begin{equation}\label{eq:decompose}
\mathbb{E}^{\pi}\left[\sum_{t \in \mathcal{T}_{j}}\left(\mu^{\ast}_{t} - \mu^{\pi}_{t}\right)\right]
 \;=\; \underbrace{\sum_{t \in \mathcal{T}_{j}}\mu^{\ast}_{t}
 - \max_{k\in \mathcal{K}}\left\{\sum_{t \in \mathcal{T}_{j}}\mu^{k}_{t}\right\}}_{J_{1,j}}
 + \underbrace{\max_{k\in \mathcal{K}}\left\{\sum_{t \in \mathcal{T}_{j}}\mu^{k}_{t}\right\}
  -  \mathbb{E}^{\pi}\left[\sum_{t \in \mathcal{T}_{j}}\mu^{\pi}_{t}\right]}_{J_{2,j}}.\vspace{-0.3cm}
\end{equation}
The first component, $J_{1,j}$, corresponds to the expected loss associated with using a single action over batch $j$. The second component, $J_{2,j}$, is the regret relative to the best static action in batch $j$.

\textbf{Step 2 (Analysis of $J_{1,j}$).} Define $\mu_{T+1}^{k} = \mu_{T}^{k}$ for all $k\in \mathcal{K}$, and denote $V_{j} = \sum_{t \in \mathcal{T}_{j}}\max_{k\in \mathcal{K}}\left|\mu^{k}_{t+1} - \mu^{k}_{t}\right|$. We note that:\vspace{-0.15cm}
\begin{equation}\label{eq:variation budg}
\sum_{j=1}^{m}V_{j}
 \;=\; \sum_{j=1}^{m}\sum_{t \in \mathcal{T}_{j}}\max_{k\in \mathcal{K}}\left|\mu^{k}_{t+1} - \mu^{k}_{t}\right|
 \;\leq\; V_{T}.\vspace{-0.15cm}
\end{equation}
Letting $k_{0}\in~\argmax_{k\in \mathcal{K}}\left\{\sum_{t \in \mathcal{T}_{j}}\mu^{k}_{t}\right\}$, we follow Step 2 in the proof of Theorem \ref{thm:resregret} to establish for any $\mu\in\mathcal{V}$ and $j\in \left\{1,\ldots,m\right\}$:\vspace{-0.2cm}
\begin{eqnarray}
J_{1,j}
&=& \sum_{t \in \mathcal{T}_{j}}\left(\mu^{\ast}_{t} - \mu^{k_0}_{t}\right)
\;\leq\; \Delta\max_{t\in \mathcal{T}_{j}} \left\{ \mu^{\ast}_{t} - \mu^{k_0}_{t}  \right\}
\;\leq\; 2V_{j}\Delta.\label{eq:upp4}\vspace{-0.2cm}
\end{eqnarray}

\textbf{Step 3 (Analysis of $J_{2,j}$.)}
We next bound $J_{2,j}$, the difference between the performance of the single best action in $\mathcal{T}_{j}$ and that of the policy, throughout $\mathcal{T}_{j}$. Let $t_{j}$ denote the first decision index of batch $j$, that is, $t_{j} = (j-1)\Delta + 1$. We denote by $W_{t}$ the sum of all weights at decision $t$: $W_{t} = \sum_{k=1}^{K}w_{t}^{k}$. Following the proof of Theorem 8.1 in \cite{Auer-etal}, one has:\vspace{-0.4cm}
\begin{equation}\label{eq:upp9}
\frac{W_{t+1}}{W_{t}} \;\leq\;
1 + \frac{\gamma / K}{1 - \gamma}X^{\pi}_{t} + \frac{(e-2)(\gamma / K)^{2}}{1 - \gamma}\sum_{k=1}^{K}\hat{X}^{k}_{t}
 + e\alpha.\vspace{-0.2cm}
\end{equation}
Taking logarithms on both sides of (\ref{eq:upp9}) and summing over all $t\in \mathcal{T}_{j}$, we get:\vspace{-0.2cm}
\begin{equation}\label{eq:upp10}
\log\left(\frac{W_{t_{j+1}}}{W_{t_{j}}}\right)
 \;\leq\; \frac{\gamma / K}{1 - \gamma}\sum_{t\in\mathcal{T}_{j}}X^{\pi}_{t}
  + \frac{(e-2)(\gamma / K)^{2}}{1 - \gamma}\sum_{t\in\mathcal{T}_{j}}\sum_{k=1}^{K}\hat{X}^{k}_{t}
 + e\alpha\left|\mathcal{T}_{j}\right|,\vspace{-0.2cm}
\end{equation}
where for $\mathcal{T}_{m}$ set $W_{t_{m+1}} = W_{T}$. Let $k_{j}$ be the best action in $\mathcal{T}_j$: $k_{j} \in \argmax_{k\in \mathcal{K}}\left\{\sum_{t \in \mathcal{T}_{j}}X^{k}_{t}\right\}$. Then,\vspace{-0.2cm}
\begin{eqnarray}
w^{k_{j}}_{t_{j+1}} &\geq& w^{k_{j}}_{t_{j}+ 1}\exp\left\{ \frac{\gamma}{K}\sum_{t_{j}+1}^{t_{j+1}-1}\hat{X}^{k_{j}}_{t}   \right\}\notag\\
&\geq& \frac{e\alpha}{K}W_{t_{j}}\exp\left\{ \frac{\gamma}{K}\sum_{t_{j}+1}^{t_{j+1}-1}\hat{X}^{k_{j}}_{t}   \right\}\notag\\
&\geq& \frac{\alpha}{K}W_{t_{j}}\exp\left\{ \frac{\gamma}{K}\sum_{t\in \mathcal{T}_{j}}\hat{X}^{k_{j}}_{t}   \right\},\notag
\end{eqnarray}
where the last inequality holds since $\gamma \hat{X}^{k_{j}}_{t}/K \leq 1$. Therefore,\vspace{-0.2cm}
\begin{equation}\label{eq:upp11}
\log\left(\frac{W_{t_{j+1}}}{W_{t_{j}}}\right)
 \geq \log\left(\frac{w^{k_{j}}_{t_{j+1}}}{W_{t_{j}}}\right)
 \geq \log\left(\frac{\alpha}{K}\right) + \frac{\gamma}{K}\sum_{t\in \mathcal{T}_{j}}X_{t}^{\pi}.\vspace{-0.2cm}
\end{equation}
Taking (\ref{eq:upp10}) and (\ref{eq:upp11}) together, one has\vspace{-0.2cm}
\[
\sum_{t\in\mathcal{T}_{j}}X^{\pi}_{t}
\;\geq\; \left(1-\gamma\right) \sum_{t\in \mathcal{T}_{j}}\hat{X}^{k_{j}}_{t}
 - \frac{K\log\left(K/\alpha\right)}{\gamma}
 - \left(e-2\right)\frac{\gamma}{K}\sum_{t\in \mathcal{T}_{j}}\sum_{k=1}^{K}\hat{X}^{k}_{t}
 - \frac{e\alpha K \left|\mathcal{T}_{j}\right|}{\gamma}.\vspace{-0.2cm}
\]
Taking expectation with respect to the noisy rewards and the actions of Exp3.S we have:\vspace{-0.25cm}
\begin{eqnarray}
J_{2,j} &=& \max_{k\in \mathcal{K}}\left\{\sum_{t \in \mathcal{T}_{j}}\mu^{k}_{t}\right\}
  - \mathbb{E}^{\pi}\left[\sum_{t \in \mathcal{T}_{j}}\mu^{\pi}_{t}\right]\notag\\
&\leq& \sum_{t \in \mathcal{T}_{j}}\mu^{k_{j}}_{t}
 + \frac{K\log\left(K/\alpha\right)}{\gamma}
 + \left(e-2\right)\frac{\gamma}{K}\sum_{t\in \mathcal{T}_{j}}\sum_{k=1}^{K}\mu^{k}_{t} + \frac{e\alpha K\left|\mathcal{T}_{j}\right|}{\gamma}
 - \left(1-\gamma\right) \sum_{t\in \mathcal{T}_{j}}\mu^{k_{j}}_{t}\notag\\
 &=& \gamma\sum_{t \in \mathcal{T}_{j}}\mu^{k_{j}}_{t}
 + \frac{K\log\left(K/\alpha\right)}{\gamma}
 + \left(e-2\right)\frac{\gamma}{K}\sum_{t\in \mathcal{T}_{j}}\sum_{k=1}^{K}\mu^{k}_{t}
 + \frac{e\alpha K \left|\mathcal{T}_{j}\right|}{\gamma}\notag\\
 &\stackrel{(a)} \leq& \left(e - 1\right)\gamma \left|\mathcal{T}_{j}\right|
 + \frac{K\log\left(K/\alpha\right)}{\gamma}
 + \frac{e\alpha K \left|\mathcal{T}_{j}\right|}{\gamma},\label{eq:compose3a}
\end{eqnarray}
for every batch $1\leq j \leq m$, where (a) holds since $\sum_{t \in \mathcal{T}_{j}}\mu^{k_{j}}_{t} \leq~\left|\mathcal{T}_{j}\right|$ and $\sum_{t\in \mathcal{T}_{j}}\sum_{k=1}^{K}\mu^{k}_{t}\leq~K\left|\mathcal{T}_{j}\right|$.

\textbf{Step 4 (Regret throughout the horizon).}
Taking (\ref{eq:upp4}) together with (\ref{eq:compose3a}), and summing over $m=\left\lceil T/\Delta\right\rceil$ batches we have:\vspace{-0.3cm}
\begin{eqnarray}
\mathcal{R}^{\pi}(\mathcal{V},T)
&\leq& \sum_{j=1}^{m}\left(\left(e - 1\right)\gamma \left|\mathcal{T}_{j}\right|
 + \frac{K\log\left(K/\alpha\right)}{\gamma}
 + \frac{e\alpha K \left|\mathcal{T}_{j}\right|}{\gamma}
 + 2V_{j}\Delta\right)\notag\\
&\leq& \left(e - 1\right)\gamma T
 + \frac{e\alpha K T}{\gamma}
 + \left( \frac{T}{\Delta} + 1\right)\frac{K\log\left(K/\alpha\right)}{\gamma}
 + 2V_{T}\Delta\notag\\
&\leq& \left(e - 1\right)\gamma T
 + \frac{e\alpha K T}{\gamma}
 + \frac{2KT\log\left(K/\alpha\right)}{\gamma\Delta}
 + 2V_{T}\Delta, \label{eq:upp13}
\end{eqnarray}
for any $\Delta\in\left\{1,\ldots,T\right\}$. Setting the tuning parameters to be $\alpha = \frac{1}{T}$ and $\gamma = \min\left\{1, \left(\frac{4V_{T}K\log\left(KT\right)}{(e-1)^{2}T}\right)^{1/3}\right\}$, and selecting a batch size
$\Delta = \left\lceil\left(\frac{e-1}{2}\right)^{1/3}
\cdot\left(K\log\left(KT\right)\right)^{1/3}
\cdot\left(\frac{T}{V_{T}}\right)^{2/3}\right\rceil$ one has:\vspace{-0.25cm}
\begin{eqnarray}
\mathcal{R}^{\pi}(\mathcal{V},T)
%\;\leq\; 3\left(2\left(e-1\right)KV_{T}\log\left(KT\right)\right)^{1/3}\cdot T^{2/3} + o\left(T^{2/3}\right).\vspace{-0.2cm}
%\;\leq\; 8(e-1)\left(KV_{T}\log\left(KT\right)\right)^{1/3}\cdot T^{2/3}.
&\leq& e\cdot\left(\frac{e-1}{2}\right)^{2/3}\frac{K^{2/3}T^{1/3}}{\left(V_{T}\log\left(KT\right)\right)^{1/3}}
\;+\; 3\cdot 2^{2/3}(e-1)^{2/3}\left(KV_{T}\log\left(KT\right)\right)^{1/3} T^{2/3}\notag\\
&\leq& \left(e\cdot\left(\frac{e-1}{2}\right)^{2/3}\;+\; 4\cdot 2^{2/3}(e-1)^{2/3}\right)\cdot\left(KV_{T}\log\left(KT\right)\right)^{1/3} T^{2/3},\notag
\end{eqnarray}
where the last inequality holds by recalling $K<T$. whenever $T$ is unknown, we can use Exp3.S as a subroutine over exponentially increasing pulls epochs $T_{\ell} = 2^{\ell}$, $\ell = 0,1,2,\ldots$, in a manner which is similar to the one described in Corollary $8.4$ in \cite{Auer-etal} to show that since for any $\ell$ the regret incurred during $T_{\ell}$ is at most $C\left(KV_{T}\log\left(KT_{\ell}\right)\right)^{1/3}\cdot~T_{\ell}^{2/3}$ (by tuning $\alpha$ and $\gamma$ according to $T_{\ell}$ in each epoch $\ell$), and for some absolute constant $\tilde{C}$, we get that $\mathcal{R}^{\pi}(\mathcal{V},T)
 \;\leq\; \tilde{C}\left(\log\left(KT\right)\right)^{1/3}\left(KV_{T}\right)^{1/3}T^{2/3}$. This concludes the proof.\qed

\small
\setstretch{1.2}
\bibliographystyle{chicago}
\bibliography{bibli}

\end{document}